\pdfoutput=1
\documentclass{article}
\usepackage{iclr2027_conference,times}

\usepackage{amsmath,amsfonts,bm}

\def\eqref#1{equation~\ref{#1}}

\def\1{\bm{1}}

\DeclareMathAlphabet{\mathsfit}{\encodingdefault}{\sfdefault}{m}{sl}
\SetMathAlphabet{\mathsfit}{bold}{\encodingdefault}{\sfdefault}{bx}{n}

\usepackage[T1]{fontenc}
\usepackage{url}

\usepackage{graphicx}
\usepackage{amsmath,amssymb}
\usepackage{booktabs}
\usepackage{xcolor}
\usepackage{multirow}
\usepackage{flafter}
\usepackage{placeins}
\usepackage{needspace}
\usepackage{float}
\usepackage{etoolbox}
\usepackage{hyperref}
\hypersetup{hidelinks}
\AtBeginEnvironment{table}{\setlength{\abovecaptionskip}{0pt}\setlength{\belowcaptionskip}{4pt}}
\newcommand{\sect}[1]{Sec.~\ref{#1}}
\newcommand{\method}{\mbox{WorldCast}}
\providecommand{\eg}{e.g.}
\newcommand{\myparagraph}[1]{%
  \noindent\textbf{#1}%
}

\makeatletter
\@ifundefined{iflatexml}{\newif\iflatexml}{}
\makeatother
\iflatexml
\newcommand{\wcauthors}{Ziyang Ye$^{1}$\quad Junchao Huang$^{1,2}$\quad Evelyn Zhang$^{2}$\quad Zhihao Xie$^{1}$\quad Ruicheng Zhang$^{3}$\quad Boyao Han$^{1}$\quad Litao Ban$^{4}$\quad Ziye Wang$^{4}$\quad Xinting Hu$^{5}$\quad Shaoshuai Shi$^{4}$\quad Zhuotao Tian$^{2}$\quad Li Jiang$^{1,2\dagger}$\\$^{1}$CUHK-Shenzhen\quad $^{2}$SLAI\quad $^{3}$Tsinghua SIGS\quad $^{4}$Voyager Research, Didi Chuxing\quad $^{5}$USTC\\$^{\dagger}$Corresponding author\quad Project page: \url{https://ziyang-ye.github.io/WorldCast-Page}}
\else
\newcommand{\wcauthors}{Ziyang Ye$^{1}$\quad Junchao Huang$^{1,2}$\quad Evelyn Zhang$^{2}$\quad Zhihao Xie$^{1}$\\
\bf Ruicheng Zhang$^{3}$\quad Boyao Han$^{1}$\quad Litao Ban$^{4}$\quad Ziye Wang$^{4}$\\
\bf Xinting Hu$^{5}$\quad Shaoshuai Shi$^{4}$\quad Zhuotao Tian$^{2}$\quad Li Jiang$^{1,2\dagger}$\\[4pt]
$^{1}$CUHK-Shenzhen\quad $^{2}$SLAI\quad $^{3}$Tsinghua SIGS\\
$^{4}$Voyager Research, Didi Chuxing\quad $^{5}$USTC\\[4pt]
Project page: \url{https://ziyang-ye.github.io/WorldCast-Page}}
\fi

\title{\method{}: Distributed Multiplayer \\ World Models}

\author{\wcauthors}

\iclrfinalcopy

\begin{document}

\hfuzz=1pt
\setlength{\floatsep}{8pt plus 2pt minus 2pt}
\setlength{\textfloatsep}{8pt plus 2pt minus 2pt}
\setlength{\intextsep}{12pt plus 1pt minus 2pt}

\maketitle
\lhead{Preprint}
\iflatexml\else
{\renewcommand{\thefootnote}{\fnsymbol{footnote}}\footnotetext[2]{Corresponding author.}}
\fi

\begin{figure}[H]
\centering
\includegraphics[trim=24.25bp 12.75bp 11.75bp 17bp,clip,width=\linewidth]{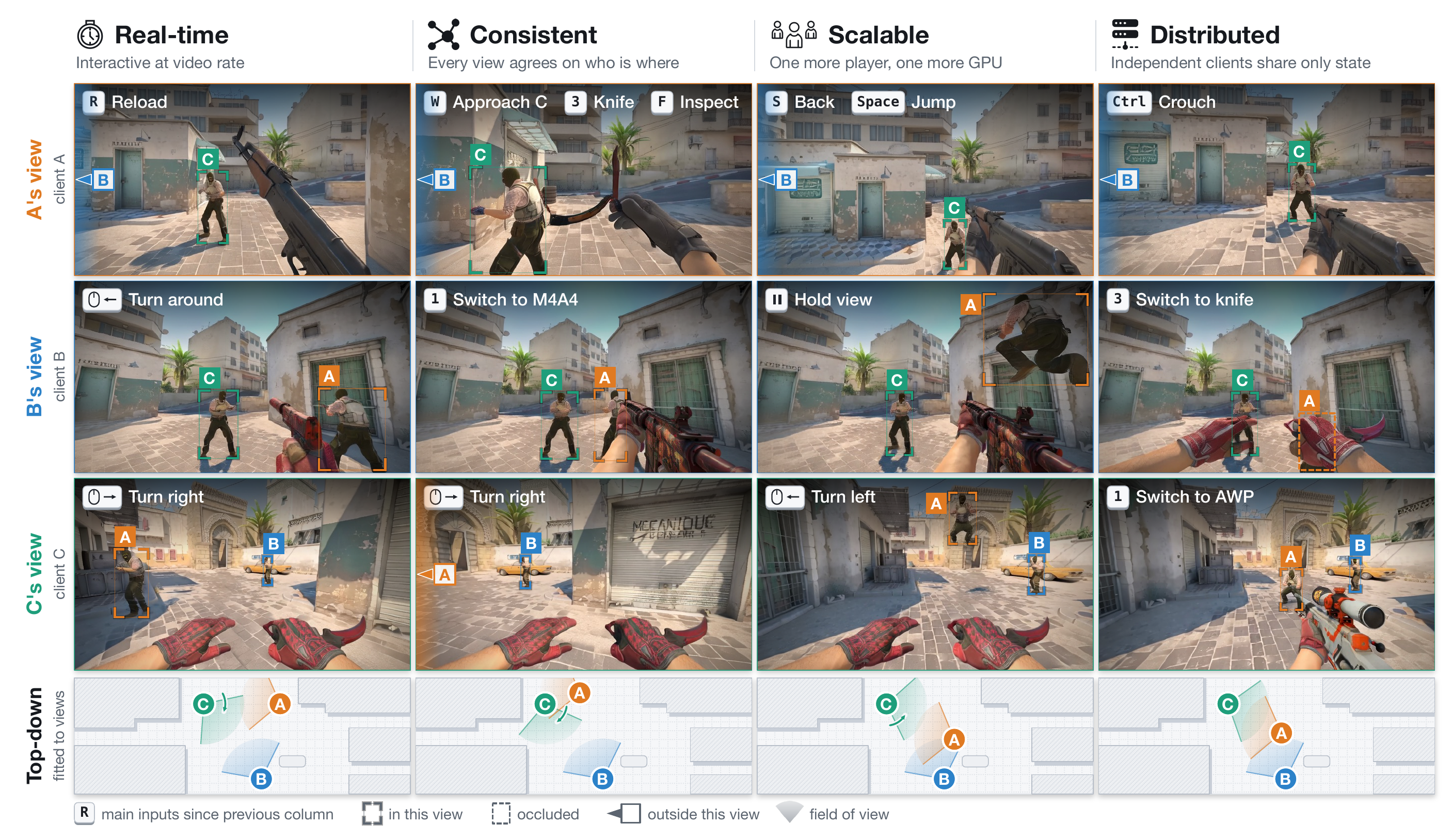}
\caption{%
Three clients, each running on a separate machine, generate independently controlled views using shared world state. Only the initial frames are provided. Columns show successive moments, with controls annotated above each view; the bottom row shows a top-down layout fitted to the generated views. WorldCast supports real-time generation, cross-view consistency, scalability by adding clients, and distributed execution through state exchange.
}
\label{fig:teaser}
\end{figure}

\begin{abstract}
Multiplayer world models must generate independently controlled views with consistent representations of both players and their shared environment.
Most existing approaches coordinate multiple players through joint multi-view generation, whose cost grows with each additional player. We present \textbf{WorldCast}, a distributed multiplayer world model in which each player runs a local client comprising a video generator and a state model. 
Using recorded player positions and map geometry during training, the state model estimates the player's position from generated video and control inputs.
Clients exchange player states and project them into camera-aligned player state fields that guide where and how other players are rendered. Shared scene state enables clients to reuse one another's generated observations to maintain consistent scene appearance across views. Experiments on Counter-Strike~2 demonstrate WorldCast's consistency, real-time performance, and distributed scalability. The camera-aligned player state field improves player rendering rates by over an order of magnitude over joint-generation methods, while shared scene state improves visual consistency over whole rounds. 
Each client runs in real time and exchanges only player and scene states, enabling scalable multiplayer generation without a centralized computational bottleneck.
Image quality remains stable over hour-long rollouts. 

\end{abstract}

\section{Introduction}
\label{sec:introduction}

Interactive video world models generate an agent's visual observations in response to its actions~\citep{bruce2024genie,skywork2025matrixgame,valevski2025gamengen,huang2026solarwmopendatascalable}. Multiplayer world modeling extends this setting to several independently controlled players in a shared environment~\citep{cai2026mass,liu2026gamma,savva2026solaris}. When players are visible, their locations should be consistent across viewpoints; previously observed parts of the environment should also retain their appearance across views and over time. Each view should remain responsive as players are added and, ideally, be generated on its own machine. Maintaining cross-view consistency without coupling the computation of different views is the central challenge.

Most prior methods generate players' views jointly through cross-view attention~\citep{hu2026mira,hu2026metaworld,liu2026gamma,mo2026worldweaver,savva2026solaris,sun2026prisma}, increasing computation and inter-device communication as players are added.
Generating views independently avoids this coupling, but each generator then has access only to its own observations and controls. It needs additional information to place other players correctly and reproduce scene content observed from other viewpoints. A distributed approach must therefore share both spatial information about players and generated scene appearance, without requiring the views to be generated together.

To provide this shared information while keeping generation independent, we propose \textbf{WorldCast}. Inspired by the organization of networked games~\citep{funkhouser1995ring,bernier2001latency}, each player runs a client that maintains a local copy of the shared world state and generates its own view.
We represent this state with two complementary components. \textbf{Player state} records players' positions and attributes, projected into a camera-aligned player state field that guides the rendering of other players. \textbf{Scene state} stores generated observations in a shared memory bank, supplying content missing from each client's recent context. Clients exchange these states once per video block while keeping intermediate features local. A fixed-resolution player state field and a bounded number of retrieved scene frames allow additional players without enlarging the generator's input.

Unlike in a networked game, WorldCast has no engine supplying positions and geometry at inference. Each client therefore includes a \textbf{state model}, trained with recorded positions and map geometry, that estimates its own position and scene depth from generated video and controls. Position estimates correct control-based extrapolation and update player state; depth supports player visibility checks and scene memory retrieval. After each block, clients publish these updates and share the generated observations, completing the loop between generation and state estimation. 

We evaluate \method{} on Counter-Strike~2 (Fig.~\ref{fig:teaser}). 
Given the same ground-truth (GT) player positions, our camera-aligned player field achieves a player rendering rate over ten times that of the joint-generation~\citep{savva2026solaris} and frame-concatenation baselines. This rate measures the proportion of visible players rendered at their expected positions.
Predicted player positions retain much of this rendering rate over ten-second windows (Table~\ref{tab:predicted}).
Shared scene state further improves visual consistency over whole rounds (Table~\ref{tab:scene}).
With scene state enabled, each client generates video at over 16\,FPS across deployments of 2 to 16 clients (Fig.~\ref{fig:deployment}), including decoding but excluding synchronization waits. Image quality remains stable over hour-long rollouts.

We summarize our contributions as follows:
\begin{enumerate}\renewcommand{\labelenumi}{\textbf{\arabic{enumi}.}}
\item We propose \method{}, a distributed multiplayer world model whose clients each generate one player's view and share only a world state, which conditions each generator through a camera-aligned player state field and a scene memory bank.
\item We develop a {state model} that estimates each player's position and scene depth from its generated video and controls, enabling clients to update the shared world state without a game engine supplying these quantities at inference.
\item We demonstrate improved player rendering and whole-round visual consistency on Counter-Strike~2. Each client generates video in real time with 2 to 16 players, and image quality remains stable over hour-long rollouts.
\end{enumerate}

\section{Related Work}
\label{sec:related}

\myparagraph{Multiplayer World Models.}
Extending single-view world models~\citep{skywork2025matrixgame,kanervisto2025wham,zhang2026robostereo}, joint-generation methods denoise all views together~\citep{hu2026mira,hu2026metaworld,liu2026gamma,mo2026worldweaver,savva2026solaris,sun2026prisma,zhu2026shareverse}, so views on different machines exchange attention features at every step (Appendix~\ref{app:deploy}); shared-screen methods bind each agent's actions to its character in one frame~\citep{pondaven2026actionparty}; shared-state methods condition views on a state advanced by actions, either centrally~\citep{cai2026mass} or by per-player generators on a given map~\citep{po2026multigen}, or encoded from all views' latest frames~\citep{wu2026multiworld}. \method{} clients exchange only player state, estimated from generated video and controls, and scene state, a memory bank of generated blocks.

\myparagraph{Controllable Video Generation.}
Structured conditioning provides spatial guidance for image and video generation. Object boxes can be encoded as additional tokens~\citep{li2023gligen,gao2023magicdrive,russell2025gaia2}, while image-aligned control maps convey edges or depth~\citep{zhang2023controlnet}, motion trajectories~\citep{wang2025ati,zhang2026zo3t} or time-varying 3D states~\citep{zheng2026versecrafter}. \method{} applies such image-aligned conditioning to incorporate the player states exchanged between clients. Its state injector projects each visible player into the local camera and places both spatial and gameplay attributes at the corresponding generator tokens.

\myparagraph{Autoregressive Video Diffusion.}
Autoregressive video diffusion models generate video incrementally in a video autoencoder's latent space~\citep{wan2025wan,xie2026videorae} with block-causal attention~\citep{chen2024diffusionforcing,yin2025causvid}. Self Forcing~\citep{huang2025selfforcing} trains on the model's own rollouts to reduce error accumulation, and LIVE~\citep{huang2026live} bounds this error to keep rollouts stable well beyond the training horizon. Distribution matching distillation~\citep{yin2024dmd2,yin2024dmd} enables few-step generators. 
To preserve scenes beyond a limited context window, memory-augmented models~\citep{huang2025memory,li2025vmem,xiao2025worldmem,yu2025contextasmemory,nam2026worldcam,sun2025worldplay} retrieve their own past frames by field-of-view overlap, camera pose or 3D geometry. ShareVerse~\citep{zhu2026shareverse} also retrieves other agents' frames by proximity within joint denoising. 
WorldCast combines four-step autoregressive generation with a memory bank shared across independent clients, allowing clients to reuse one another's generated content without jointly denoising their views.

\section{Method}
\label{sec:method}

WorldCast lets multiple players inhabit one generated world, kept consistent through a shared world state. This state comprises player state, which describes players' positions and attributes, and scene state, which stores generated observations of the environment. We first introduce the distributed generation framework (Sec.~\ref{sec:framework}), then describe how player state (Sec.~\ref{sec:field}) and scene state (Sec.~\ref{sec:scene}) are maintained and incorporated into each generator. We next present the state model that estimates position and depth from generated video (Sec.~\ref{sec:grounding}), followed by training and inference (Sec.~\ref{sec:streaming}).

\begin{figure}[!t]
\centering
\includegraphics[trim=12bp 11.5bp 11.75bp 12bp,clip,width=\linewidth]{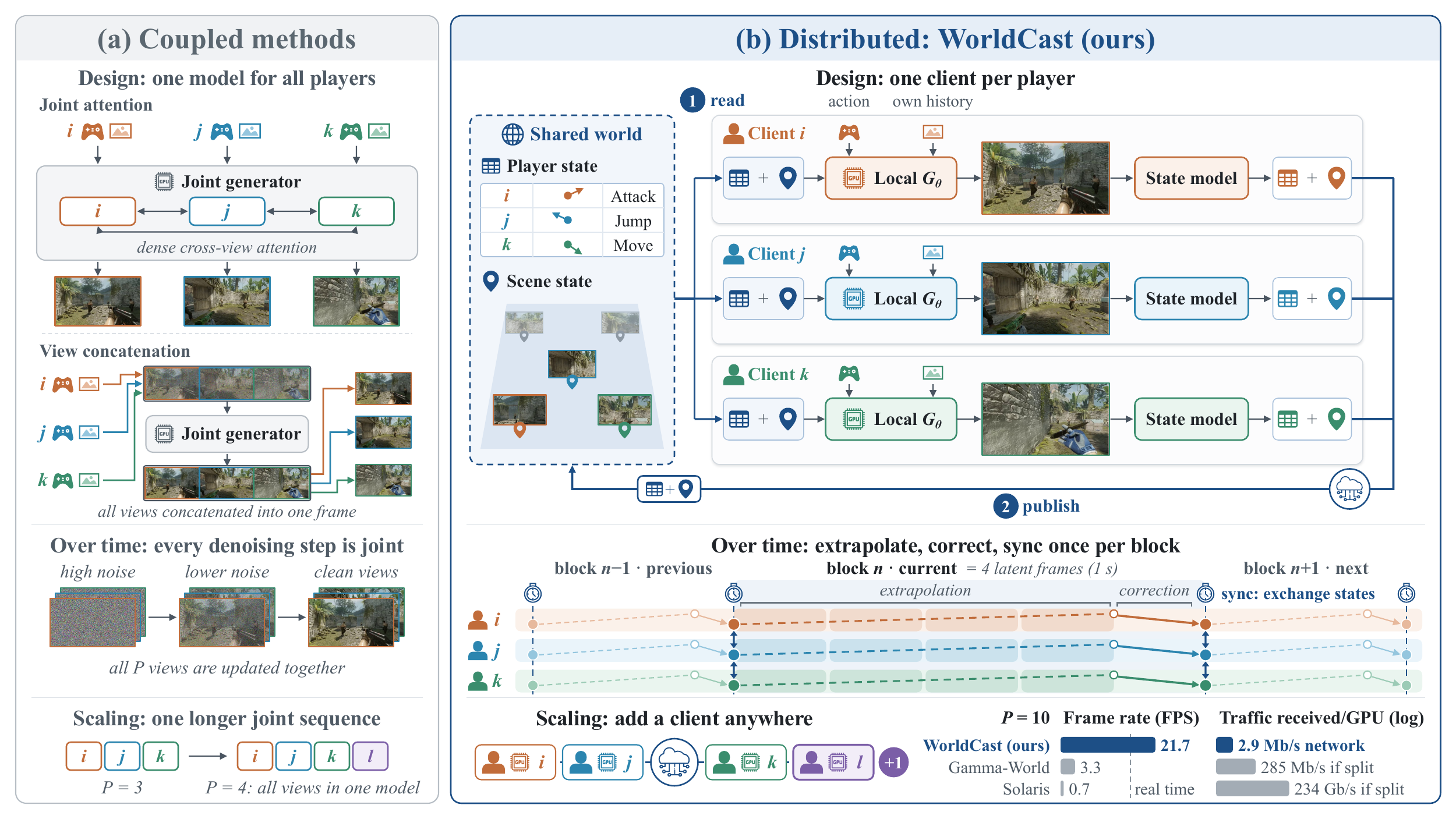}
\caption{\textbf{(a)} Coupled designs generate all players' views in one model. \textbf{(b)} Each \method{} player runs its own client, and clients exchange only the shared world state.}
\label{fig:architecture}
\end{figure}

\subsection{Distributed Generation with Shared World State}
\label{sec:framework}

To support scalable multiplayer video generation across machines, WorldCast separates shared world state from local rendering. As in Fig.~\ref{fig:architecture}(a), existing coupled designs coordinate players 
through joint multi-view generation that processes players' views together through cross-view attention or view concatenation. 
WorldCast instead assigns each of the $P$ players an independent client that reads shared world state, generates its own view, and publishes updated state once per block (Fig.~\ref{fig:architecture}(b)). 
Each client runs on its own GPU and combines a video generator $G_\theta$ with a state model that estimates position and scene depth from generated video and controls. Only the shared world state is exchanged; activations, KV caches, and intermediate features remain local.

Consistent views require information beyond each client's recent context. The shared world state therefore comprises two complementary components: \textbf{player state}, a record of each player's position and attributes (Sec.~\ref{sec:field}), and \textbf{scene state}, a memory bank shared by all clients (Sec.~\ref{sec:scene}). Player state tracks players outside the local view, while scene state preserves observations no longer in the recent context or available only from other clients. Together, these shared states help clients render players and scene appearance consistently across independently generated views.

Clients exchange state once per block and, in our evaluations, advance in lockstep (Appendix~\ref{app:deploy}). For generating block $n$, client $i$ uses the shared states published after block $n-1$, extrapolating player positions and camera poses using the player controls. It then constructs a camera-aligned player state field and retrieves at most one memory entry from the memory bank. Specifically, for client $i$ at block $n$, the generation is conditioned as
\begin{equation}
\mathbf{V}^i_n = G_\theta\bigl(\mathbf{a}^i_n,\; \mathbf{F}^i_n,\; \mathbf{M}^i_n,\; \mathbf{V}^i_{<n}\bigr),
\label{eq:gen}
\end{equation}
where $\mathbf{a}^i_n$ denotes the player's controls and attributes (\eg, weapon);
$\mathbf{F}^i_n=\{\mathbf{F}^i_{n,f}\}_f$ is the player state field, formed by projecting other players' states onto client $i$'s camera-aligned token grid for each latent frame $f$ (Sec.~\ref{sec:field});
$\mathbf{M}^i_n$ contains the latent frames of the retrieved scene memory entry; and $\mathbf{V}^i_{<n}$ is the recent context. After generation, the state model (Sec.~\ref{sec:grounding}) estimates the player's position and scene depth from the generated latents and controls. The client publishes its updated player state and shares the generated block with its cameras and depth as a new memory entry.

The player state field has a fixed spatial resolution, and each generator reads at most one memory entry, regardless of the player count. Additional players can therefore be served by adding clients without enlarging each generator's attention sequence.

\subsection{Shared Player State}
\label{sec:field}

\myparagraph{Attributes and update.}
The shared player state contains one record $\mathbf{s}^i_n$ for each player $i$ at block $n$, describing its position and other attributes, such as facing direction, controls, and weapon. Each record is updated and published only by that player's client. The position is estimated from generated video and controls by the state model (Sec.~\ref{sec:grounding}), while other attributes are obtained from the player's inputs. These records provide a common reference for rendering players across clients. 

\myparagraph{Camera-aligned player state field.} 
Player states are expressed in world coordinates, whereas the generator operates on image tokens. The state injector bridges these representations by projecting other players into the observing client's camera and encoding their attributes at the corresponding token locations. 
Specifically, for each latent frame $f$, it constructs a player state field $\mathbf{F}_{n,f}^i$ on the generator's token grid. 
Each player within the view frustum and visible (Sec.~\ref{sec:grounding}) is projected into the observer's camera and splatted with a Gaussian kernel~\citep{wang2025ati}. The field encodes coverage, camera-relative depth and facing direction, and gameplay attributes such as team, identity, weapon, and controls. Contributions from overlapping players are combined according to their spatial weights and depths (Appendix~\ref{app:field}). This gives the generator explicit spatial guidance about where other players should appear and which attributes to render.

\myparagraph{Injection into the generator.}
A convolutional stem $\mathrm{Conv}_0$ with a zero-initialized output projection~\citep{zhang2023controlnet} maps each $\mathbf{F}_{n,f}^i$ to the model width and adds it after an early DiT block:
\begin{equation}
\mathbf{h}_{f,u} \leftarrow \mathbf{h}_{f,u} + \mathrm{Conv}_0(\mathbf{F}_{n,f}^i)(u),
\label{eq:write}
\end{equation}
where $\mathbf{h}_{f,u}$ is the hidden state of token $u$ in latent frame $f$.

\subsection{Shared Scene State}
\label{sec:scene}

\myparagraph{Scene state composition.} 
Shared scene state allows clients to reuse generated observations of the environment, helping maintain consistent scene appearance across views. It is represented as a memory bank, where each entry stores a block of latent frames generated by one client, together with their camera poses and coarse depth maps estimated by the state model (Sec.~\ref{sec:grounding}). 

\myparagraph{Scene memory update.}
Each client publishes every generated block as a new memory entry and maintains at most $B$ of its own entries. When this capacity is exceeded, it withdraws the most redundant entry---the one whose content is best covered by the client's other entries under reprojection (Appendix~\ref{app:scene}). Each entry therefore has a single owner responsible for its publication and removal, while remaining accessible to all clients. 

\myparagraph{Scene memory retrieval.} 
Before generating block $n$, client $i$ retrieves the memory entry $\mathbf{M}_n^i$ that best covers scene content missing from its recent context. It first back-projects the recent latent frames into 3D and projects these points into the cameras of the upcoming block. Pixels reached by none of these points are treated as missing. The client then projects the points of all memory entries into the same cameras. At each missing pixel, the nearest projected point to the camera defines the reference surface depth. An entry covers that pixel if one of its points lies within a depth tolerance of this surface, preventing occluded content from contributing to its coverage score. The client retrieves the entry covering the most missing pixels, or none if no entry covers any.

\begin{figure}[!t]
\centering
\includegraphics[trim=0bp 0bp 0bp 0bp,clip,width=\linewidth]{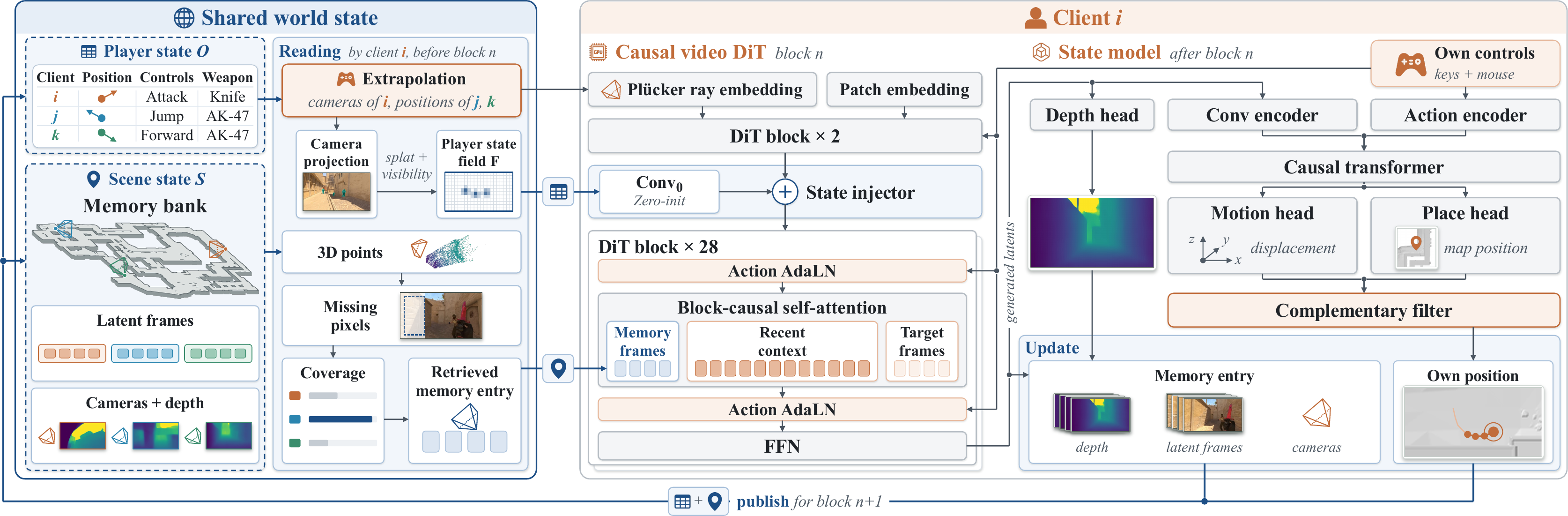}
\caption{A client and the shared world state. \textbf{Left:} player states are projected into the camera-aligned player state field, and each generated block is added to the memory bank of the scene state, from which the next block retrieves one memory entry. \textbf{Right:} the causal video DiT of client $i$, where the player state field enters after the second DiT block, the memory frames join the recent and target frames, and every frame carries Pl\"{u}cker rays; after block $n$, the state model estimates the player's position and the depth of each generated frame, which are published for block $n+1$.}
\label{fig:scene-state}
\end{figure}

\myparagraph{Conditioning the generator.}
As shown in Fig.~\ref{fig:scene-state} right, the latent frames of the retrieved memory entry $\mathbf{M}_n^i$ are prepended to the recent context frames from $\mathbf{V}^i_{<n}$ and the target frames $\mathbf{V}^i_n$. Memory frames attend only to memory frames, while recent and target frames can attend to them under the generator's block-causal attention mask. Besides, to express the geometry of observations from different viewpoints, we encode each token's viewing ray using Plücker coordinates relative to the first target frame's camera:
\begin{equation}
\mathbf{r}_{f,u} = \bigl(\mathbf{d}_{f,u},\; (\mathbf{o}_f/\ell)\times\mathbf{d}_{f,u}\bigr)\in\mathbb{R}^6,
\qquad
\mathbf{h}_{f,u} \leftarrow \mathbf{h}_{f,u} + \phi(\mathbf{r}_{f,u}),
\label{eq:rays}
\end{equation}
where $\mathbf{d}_{f,u}$ is the unit viewing direction of token $u$ in latent frame $f$, $\mathbf{o}_f$ is the camera center, and $\ell$ is a fixed length scale. The MLP $\phi$ maps these coordinates to an embedding added before the first DiT block. This enables the generator to use shared scene observations together with their relative camera geometry through its existing attention layers.

\subsection{The State Model}
\label{sec:grounding}
\label{sec:row}
\label{sec:visibility}

The shared player and scene states require positions and geometry that correspond to the generated world. At inference, these quantities are not supplied by a game engine. Each client therefore uses a state model to estimate its own position and scene depth from generated latents and controls (Fig.~\ref{fig:scene-state}, right). Position estimates update the shared player state, while depth estimates support scene memory retrieval and player visibility.

\myparagraph{State model architecture.}
As shown in Fig.~\ref{fig:scene-state} right, a convolutional encoder maps each latent frame to a token, an action encoder adds the corresponding control embedding, and a causal transformer processes the resulting sequence. 
A motion head predicts the player's displacement between consecutive latent frames. A place head provides an absolute position estimate on the current map by classifying the frame into a map cell and predicting its offset within the cell (Appendix~\ref{app:state}). A separate depth head predicts a coarse depth map from the generated latents.

\myparagraph{Position estimation.}
The motion head captures displacement from changes in the generated frames, including stops at obstacles, but integrating its displacements accumulates error. The place head provides a map-based reference, including height, but can occasionally mis-localize a frame. We combine their estimates with a complementary filter:
\begin{equation}
\mathbf{p}_f = \tfrac12\,(\mathbf{p}_{f-1} + \boldsymbol{\Delta}_f) + \tfrac12\,\mathbf{A}_f,
\label{eq:position}
\end{equation}
where $\mathbf{p}_f$ is the position estimate at latent frame $f$, $\boldsymbol{\Delta}_f$ is the motion-head displacement, and $\mathbf{A}_f$ is the place-head estimate, which is the offset to start at the given initial position $\mathbf{p}_0$.
After each completed block, the client publishes the position estimates, together with its other attributes, as the updated player state.
Before generating the next block, each client extrapolates its camera poses and other players' positions from their latest published player states by integrating movement and view controls, as networked games do between server updates~\citep{bernier2001latency}. These predictions guide player state field construction and scene memory retrieval. Since extrapolation ignores collisions and terrain height, it may diverge from the motion depicted in the generated video. Therefore, once the block is complete, the state model uses the new frames and Eq.~(\ref{eq:position}) to correct the client's own position for the next state update.

\myparagraph{Depth and visibility.}
The predicted depth maps and camera poses back-project generated observations into 3D points in a common map coordinate system. These points support both scene memory retrieval (Sec.~\ref{sec:scene}) and visibility gating for the player state field (Sec.~\ref{sec:field}). 
Before generation, the client projects points from its recent context into the cameras of the upcoming block. Their depths determine whether other players are occluded: a player is considered visible when context points 
reach its projection 
and their depth lies no more than a fixed margin in front of it (Appendix~\ref{app:field}). Since the context may not cover newly exposed regions, the depth head processes the clean estimate $\hat{\mathbf{x}}_0$ after the first denoising step. This depth updates visibility for the remaining steps.

\myparagraph{State model training.}
The motion and place heads are supervised with recorded player positions, and the depth head is supervised with depth maps rendered from the game's public 3D map assets. These supervision targets are used only in training (Appendix~\ref{app:state}).

\subsection{Training and Inference}
\label{sec:streaming}

\myparagraph{Training stages.}
We train $G_\theta$ in four stages (Appendix Fig.~\ref{fig:stages}). Stage 1 fine-tunes a pretrained video diffusion model on Counter-Strike~2 with bidirectional attention on short clips. 
Stage 2 conditions $G_\theta$ on the shared world state, adding the player state field, then scene state.
Player states are GT states taken from recorded matches, and visibility labels are computed from the game's public 3D map assets. Memory frames are GT frames of one block from another player's view in the same round, selected to cover what the recent context misses (Appendix~\ref{app:scene}).
Stages 3 and 4 make generation real-time: stage 3 converts $G_\theta$ to block-causal attention~\citep{chen2024diffusionforcing,yin2025causvid} through teacher forcing, and stage 4 distills it to a four-step generator by distribution matching~\citep{yin2024dmd2,yin2024dmd}, trained on its own rollouts to reduce accumulated error~\citep{huang2025selfforcing}.

\myparagraph{Objective.}
Stages 1 to 3 minimize the weighted flow-matching loss:
\begin{equation}
\mathcal{L} = \mathbb{E}_{t,\boldsymbol{\epsilon}}\Bigl[\,\sum_k \alpha_k\, c_k\, \bigl\lVert \mathbf{v}_\theta(\mathbf{x}_t,t)_k - \mathbf{v}_k \bigr\rVert_2^2 \Bigr],
\label{eq:loss}
\end{equation}
where $\mathbf{x}_t$ is the latent noised to level $t$ with noise $\boldsymbol{\epsilon}$, $k$ indexes latent positions, and $\mathbf{v}_\theta$ predicts the target velocity $\mathbf{v}$ using the conditioning in Eq. (\ref{eq:gen}). Two weights emphasize regions that a uniform loss would underweight. From stage 2 onward, the foreground weight $\alpha_k$ increases the loss at positions occupied by visible players, with larger weights for smaller projected players; elsewhere, $\alpha_k=1$. For memory-conditioned samples, only target frames are supervised. The memory weight $c_k$ emphasizes target regions covered by the memory frames but absent from the recent context, and is normalized to have unit mean within each target frame. Without memory frames, $c_k=1$.

\myparagraph{Inference.}
At inference, each client executes the generation cycle of Sec.~\ref{sec:framework} with the four-step generator. 
The completed block is decoded, and the client updates and publishes its player state and a new memory entry for subsequent generation.

\section{Experiments}
\label{sec:experiments}
\label{sec:criterion}

\textbf{Datasets.} On four Counter-Strike~2 maps of OpenCS2~\citep{blanchon2026opencs2}, split by match into 303 training and 40 held-out matches, we evaluate on 256 ten-second windows and on 32 whole rounds generated concurrently by three clients of one team (five in Table~\ref{tab:longhorizon}a), the other players following the recording. 
Unless marked as predicted, methods receive GT player positions from the recording and visibility from map geometry, so that errors of the state model do not confound the comparison.

\textbf{Metrics.} 
We report \emph{FVD}, \emph{PSNR}, \emph{SSIM}, and \emph{LPIPS} for video quality. Using YOLO11~\citep{ultralytics2024yolo11}, \emph{Rendered} (the rendering rate) measures the fraction of visible players detected in the recording that are rendered within one body width of their state-specified positions. \emph{Unmatched} measures the fraction of generated detections without a nearby match to a visible player or a recorded detection. Both are medians over views of ten-second windows (Appendix~\ref{app:longhorizon}). 
Whole-round rollouts drift from the recording, so we measure consistency between two clients with overlapping views at the same time (Appendix~\ref{app:scene}). \emph{Mat.\ Pix.}~\citep{bai2025syncammaster} counts pixels matched by GIM~\citep{shen2024gim}. \emph{Inliers} counts LoFTR~\citep{sun2021loftr} matches consistent with one relative pose, and \emph{RotErr}~\citep{he2025cameractrl} is its rotation error. \emph{Depth} is the fraction of pixels where both clients' predicted depths agree. \emph{VBench} measures visual quality over hour-long rollouts.

\textbf{Implementation.} $G_\theta$ is initialized from Wan2.2-TI2V-5B~\citep{wan2025wan}; a block is 4 latent frames (1\,s at 16\,fps), and the recent context holds 12 frames.

\subsection{Main Results}
\label{sec:rendering}

\begin{figure}[!tb]
\centering
\includegraphics[trim=17.75bp 23bp 15.25bp 0.5bp,clip,width=\linewidth]{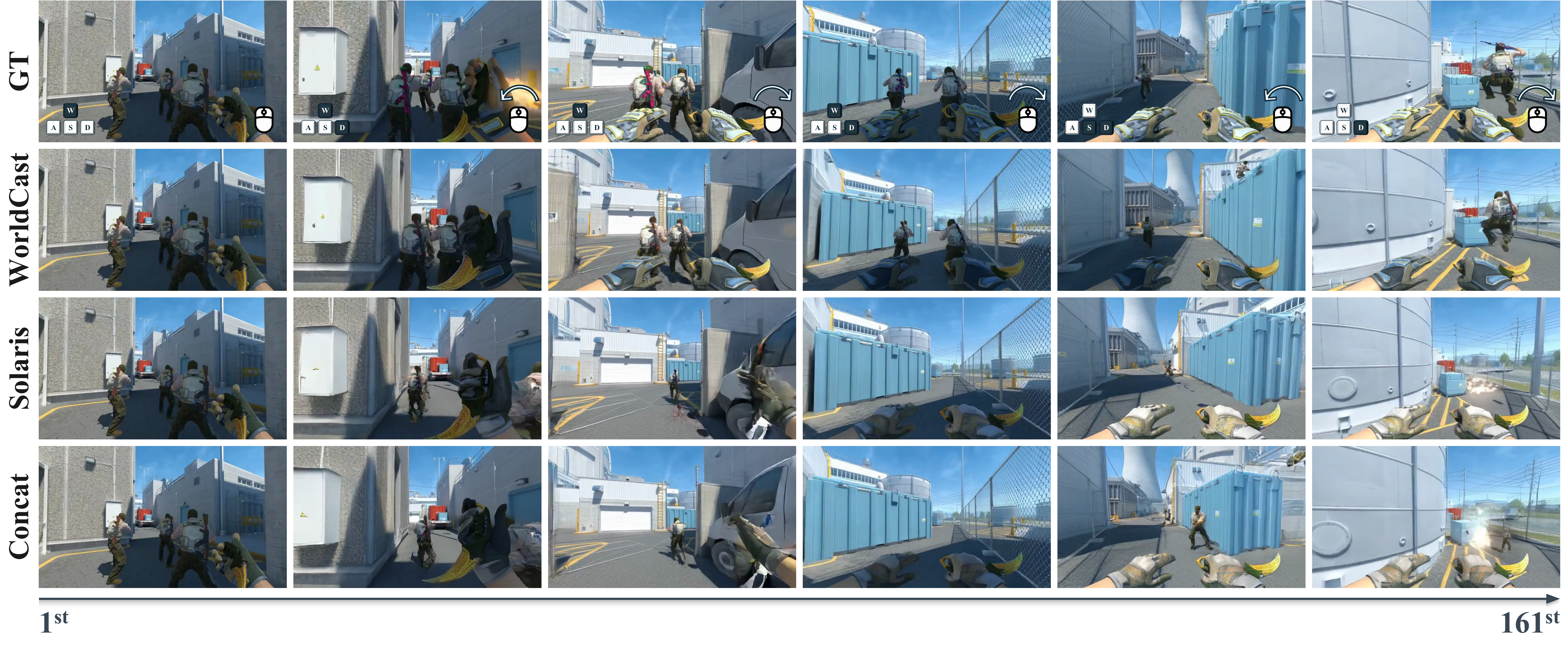}
\caption{Ten seconds of a held-out round, generated from the first frame and GT player positions (GT: recording with controls; Concat: frame concatenation).}
\label{fig:qualitative}
\end{figure}

\textbf{Bidirectional Model Results and Comparisons.} 
We compare against Solaris~\citep{savva2026solaris}, which uses joint cross-view attention, and frame concatenation. For a controlled comparison, we retrain both with the same backbone, training data, budget, and GT player positions. Our bidirectional model uses only shared player state in this comparison. As shown in Table~\ref{tab:main}, WorldCast outperforms both baselines across all reported metrics. Its player rendering rate reaches 0.817, compared with 0.056 for Solaris and 0.074 for frame concatenation, while its unmatched rate falls to 0.039. These gains demonstrate more accurate player placement alongside improved video quality (also shown in Fig.~\ref{fig:qualitative}).

\textbf{AR Model Results and State Model in the Loop.}
To test whether the state model can replace GT player positions, every client renders the other players at the positions and visibility predicted by the state models (Table~\ref{tab:predicted}). The render rate stays close to that with GT player positions, with most visible players still drawn as expected, although the unmatched rate rises from 0.031 to 0.093.

\begin{table}[!tb]
\caption{
Bidirectional multiplayer generation on ten-second windows. All methods use the same backbone, training data, budget, and GT player positions.
}
\label{tab:main}
\centering
\footnotesize
\setlength{\tabcolsep}{4pt}
\begin{tabular*}{\linewidth}{@{\extracolsep{\fill}}lcccccc@{}}
\toprule
& FVD $\downarrow$ & PSNR $\uparrow$ & SSIM $\uparrow$ & LPIPS $\downarrow$ & Rendered $\uparrow$ & Unmatched $\downarrow$ \\
\midrule
Frame concatenation & 48.2 & 15.37 & 0.493 & 0.460 & 0.074 & 0.771 \\
Solaris & 49.0 & 15.66 & 0.501 & 0.449 & 0.056 & 0.739 \\
\method{} & \textbf{39.7} & \textbf{16.17} & \textbf{0.507} & \textbf{0.418} & \textbf{0.817} & \textbf{0.039} \\
\bottomrule
\end{tabular*}

\vspace{3pt}
\caption{
Four-step AR generation results on ten-second windows and effect of predicted states. Both settings use shared player state and scene state. Joint-generation methods have no AR version.
}
\label{tab:predicted}
\begin{tabular*}{\linewidth}{@{\extracolsep{\fill}}lcccccc@{}}
\toprule
& FVD $\downarrow$ & PSNR $\uparrow$ & SSIM $\uparrow$ & LPIPS $\downarrow$ & Rendered $\uparrow$ & Unmatched $\downarrow$ \\
\midrule
WorldCast (GT positions) & 47.5 & 16.03 & 0.497 & 0.423 & 0.897 & 0.031 \\
WorldCast (Predicted positions) & 53.0 & 15.65 & 0.490 & 0.443 & 0.830 & 0.093 \\
\bottomrule
\end{tabular*}
\end{table}

\textbf{Scene State over Whole Rounds.}
\label{sec:scene-evaluation}
\method{} improves PSNR, SSIM and LPIPS after 30\,s and over whole rounds (Table~\ref{tab:scene}) over every design choice except GT frames in the memory bank, which no client has; before 30\,s, the model trained without scene state is on par (Table~\ref{tab:scene}).

\begin{table}[!tb]
\caption{Scene state over whole rounds. We use GT player positions to exclude the impact of imprecise player position in rendering performance.
}
\label{tab:scene}
\centering
\footnotesize
\setlength{\tabcolsep}{3pt}
\begin{tabular*}{\linewidth}{@{}l@{\extracolsep{\fill}}cccccc@{}}
\toprule
& \multicolumn{3}{c}{First 30\,s} & \multicolumn{3}{c}{After 30\,s} \\
\cmidrule(lr){2-4} \cmidrule(l){5-7}
& PSNR\,$\uparrow$ & SSIM\,$\uparrow$ & LPIPS\,$\downarrow$ & PSNR\,$\uparrow$ & SSIM\,$\uparrow$ & LPIPS\,$\downarrow$ \\
\midrule
Trained without scene state & \textbf{14.93} & \textbf{0.439} & \textbf{0.507} & 12.96 & 0.387 & 0.603 \\
Scene state off at inference & 14.37 & 0.424 & 0.545 & 12.93 & 0.391 & 0.615 \\
Only its own earlier blocks & 14.56 & 0.425 & 0.539 & 13.53 & 0.399 & 0.591 \\
Least-covering memory & 14.27 & 0.421 & 0.550 & 12.70 & 0.380 & 0.628 \\
\method{} & 14.88 & 0.431 & 0.521 & \textbf{14.16} & \textbf{0.409} & \textbf{0.556} \\
\midrule
GT frames in the memory bank & 15.17 & 0.442 & 0.504 & 14.91 & 0.428 & 0.517 \\
\bottomrule
\end{tabular*}
\end{table}

\begin{figure}[!tb]
\centering
\includegraphics[trim=0bp 0.25bp 0bp 1.75bp,clip,width=\linewidth]{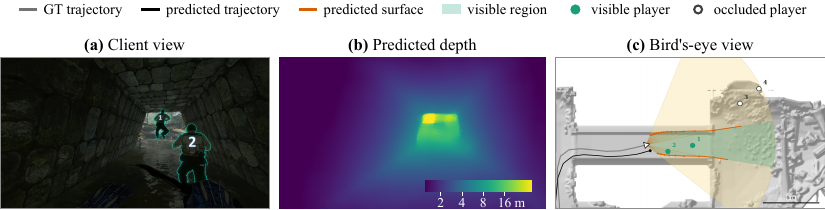}
\caption{State model on a recorded frame: \textbf{(a)} frame, \textbf{(b)} predicted depth, \textbf{(c)} predicted and GT trajectories and visibility over the map.}
\label{fig:depth}
\end{figure}

\textbf{State Estimation.}
\label{sec:state-evaluation}
On rollout frames, the state model estimates positions more accurately than integrating the controls and keeps its error low over a minute (Appendix Table~\ref{tab:state}; Fig.~\ref{fig:depth}).

\begin{figure}[!tb]
\centering
\includegraphics[trim=5bp 2.5bp 4.25bp 8.5bp,clip,width=\linewidth]{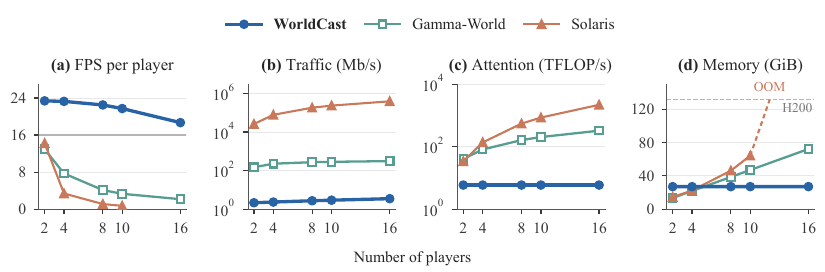}
\caption{Serving $P$ players on H200 GPUs: \textbf{(a)} FPS per player with decoding, \textbf{(b)} traffic received, \textbf{(c)} attention compute, \textbf{(d)} memory per GPU; Gamma-World and Solaris run on one GPU as released ((b): one player per GPU).}
\label{fig:deployment}
\end{figure}

\begin{figure}[!tb]
\centering
\includegraphics[trim=17.75bp 22bp 20.25bp 0.5bp,clip,width=\linewidth]{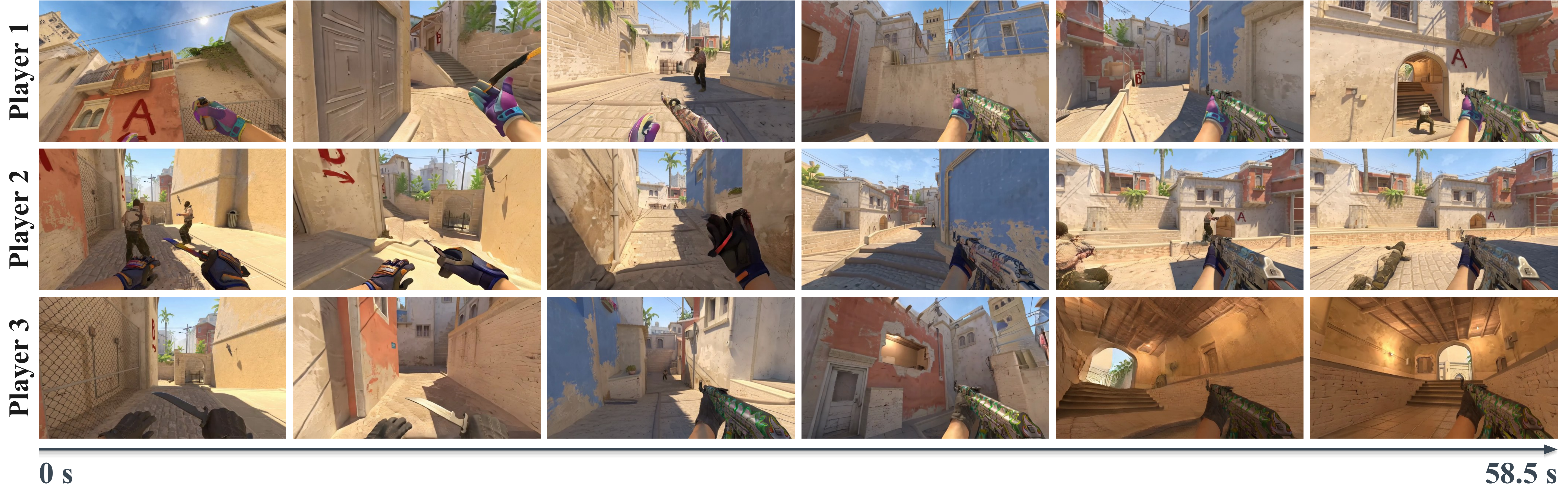}
\caption{Closed-loop deployment: a whole round generated concurrently, each client on its own GPU and starting from its first recorded frame.}
\label{fig:trio}
\end{figure}

\textbf{Closed-Loop Deployment.}
\label{sec:deployment}
\label{sec:longhorizon}
Closed-loop deployment gives each client only its recorded first frame and position (Fig.~\ref{fig:trio}). After each block, it publishes its estimated position, renders other players at their published positions, and adds the block to the scene state. The scene state keeps clients more consistent with each other than with it off at inference: all four agreement measures improve (Table~\ref{tab:longhorizon}a). In one-hour rollouts, VBench scores change little (Table~\ref{tab:longhorizon}b). With one GPU per player, each client exceeds 16\,FPS for 2 to 16 players and receives only the others' player states and memory entries (a few Mb/s), while Gamma-World~\citep{liu2026gamma} and Solaris, smaller and hosting all players on one GPU as released, fall below 16\,FPS from two players (Fig.~\ref{fig:deployment}).

\begin{table}[!tb]
\caption{\textbf{(a)} Closed-loop deployment over whole rounds, scene state off at inference and on: agreement between clients. \textbf{(b)} VBench~\citep{huang2024vbench} over one-hour rollouts.}
\label{tab:longhorizon}
\centering
\footnotesize
\setlength{\tabcolsep}{3pt}
\begin{minipage}[t]{.46\linewidth}
\vspace{0pt}\centering
\begin{tabular*}{\linewidth}{@{\extracolsep{\fill}}lcccc@{}}
\toprule
& \multicolumn{2}{c}{First 30\,s} & \multicolumn{2}{c}{After 30\,s} \\
\cmidrule(lr){2-3} \cmidrule(l){4-5}
\textbf{(a)} Scene state & Off & On & Off & On \\
\midrule
Mat.\ Pix.\ (K) $\uparrow$ & 72.0 & \textbf{92.1} & 31.1 & \textbf{44.3} \\
Inliers $\uparrow$ & 46.2 & \textbf{51.5} & 18.8 & \textbf{27.7} \\
RotErr ($^\circ$) $\downarrow$ & 31.8 & \textbf{29.0} & 59.1 & \textbf{55.6} \\
Depth $\uparrow$ & 0.243 & \textbf{0.247} & 0.092 & \textbf{0.120} \\
\bottomrule
\end{tabular*}
\end{minipage}\hfill
\begin{minipage}[t]{.50\linewidth}
\vspace{0pt}\centering
\begin{tabular*}{\linewidth}{@{\extracolsep{\fill}}lccc@{}}
\toprule
& \multicolumn{3}{c}{Minutes into the hour} \\
\cmidrule(l){2-4}
\textbf{(b)} VBench & 0--20 & 20--40 & 40--60 \\
\midrule
Subject consistency $\uparrow$ & 0.803 & 0.797 & 0.797 \\
Background consistency $\uparrow$ & 0.910 & 0.912 & 0.910 \\
Imaging quality $\uparrow$ & 70.3 & 70.3 & 69.2 \\
Aesthetic quality $\uparrow$ & 0.486 & 0.489 & 0.486 \\
\bottomrule
\end{tabular*}
\end{minipage}
\end{table}

\begin{table}[!tb]
\caption{Ablations of the player state field (stage 2, ten-second windows, 6{,}000 training steps each).}
\label{tab:ablations}
\centering
\small
\setlength{\tabcolsep}{4pt}
\begin{tabular*}{\linewidth}{@{\extracolsep{\fill}}lcccccc@{}}
\toprule
& FVD $\downarrow$ & PSNR $\uparrow$ & SSIM $\uparrow$ & LPIPS $\downarrow$ & Rendered $\uparrow$ & Unmatched $\downarrow$ \\
\midrule
\method{}      & 45.7 & 15.54 & 0.495 & 0.455 & \textbf{0.763} & 0.076 \\
Trained without camera alignment & 72.2 & 15.17 & 0.490 & 0.479 & 0.093 & 0.826 \\
Trained without visibility & 45.8 & 15.67 & 0.498 & 0.449 & 0.667 & 0.088 \\
Trained without foreground weight & 43.9 & 15.71 & 0.500 & 0.446 & 0.691 & \textbf{0.060} \\
Trained with late injection & 47.2 & 15.52 & 0.499 & 0.456 & 0.549 & 0.203 \\
Trained with a coarse field & 41.9 & 15.69 & 0.497 & 0.445 & 0.708 & 0.083 \\
Field off at inference & 73.2 & 15.34 & 0.491 & 0.467 & 0.000 & 0.333 \\
Visibility off at inference & 60.8 & 15.15 & 0.486 & 0.475 & 0.748 & 0.415 \\
\bottomrule
\end{tabular*}
\end{table}

\subsection{Ablation Studies}
\label{sec:ablations}

In the player state field (Table~\ref{tab:ablations}), camera alignment matters most: without it, the render rate falls to that of joint generation, and every other retrained variant also lowers it; disabling the visibility test at inference draws occluded players and raises the unmatched rate. For the scene state, retrieving any client's best-covering entry (our method) achieves the best results after 30\,s (Table~\ref{tab:scene}).

\section{Conclusion and Limitations}
\label{sec:conclusion}

\method{} explores a world model as the engine of a networked game, in which each client generates one player's view on its own GPU and clients exchange only the shared world state. Our method attains a player rendering rate over an order of magnitude higher than joint generation, and scene state improves whole-round consistency. The main limitation is the state model's position error on generated video, which grows with time (Appendix~\ref{app:lim}).

\FloatBarrier
\bibliographystyle{iclr2027_conference}
\bibliography{references}

\clearpage
\appendix
\makeatletter
\setlength{\@fptop}{0pt}\setlength{\@fpsep}{12pt plus 2pt}\setlength{\@fpbot}{0pt plus 1fil}
\makeatother
\setcounter{topnumber}{4}\setcounter{bottomnumber}{2}\setcounter{totalnumber}{6}
\renewcommand{\bottomfraction}{0.6}\renewcommand{\floatpagefraction}{0.75}
\section*{Appendix contents}
\noindent\begin{tabular*}{\linewidth}{@{}l@{\extracolsep{\fill}}r@{}}
\ref{app:state}\hspace{0.6em}\hyperref[app:state]{State model in detail} & \pageref{app:state} \\
\ref{app:deploy}\hspace{0.6em}\hyperref[app:deploy]{Distributed deployment in detail} & \pageref{app:deploy} \\
\ref{app:scaling}\hspace{0.6em}\hyperref[app:scaling]{Training stages and scaling} & \pageref{app:scaling} \\
\ref{app:longhorizon}\hspace{0.6em}\hyperref[app:longhorizon]{Evaluation protocols and long horizons in detail} & \pageref{app:longhorizon} \\
\ref{app:scene}\hspace{0.6em}\hyperref[app:scene]{Scene state in detail} & \pageref{app:scene} \\
\ref{app:field}\hspace{0.6em}\hyperref[app:field]{The player state field in detail} & \pageref{app:field} \\
\ref{app:lim}\hspace{0.6em}\hyperref[app:lim]{Limitations and scope} & \pageref{app:lim} \\
\ref{app:qualitative}\hspace{0.6em}\hyperref[app:qualitative]{Additional qualitative comparisons} & \pageref{app:qualitative} \\
\end{tabular*}
\vspace{1.5\baselineskip}

\noindent Throughout the appendix, \emph{GT states} are the recorded player states, whose positions are the GT player positions of \sect{sec:experiments}, with visibility from the map geometry; \emph{predicted states} replace these positions and the visibility with those predicted by the state model.

\section{State model in detail}
\label{app:state}
This appendix details the state model of \sect{sec:grounding}. Its \emph{motion head} predicts the player's displacement between consecutive latent frames, its \emph{place head} recognizes where on the map a frame was taken, and the complementary filter of Eq.~(\ref{eq:position}) advances one estimate by the former and corrects it with the latter.

\paragraph{Components.}
The state model reads windows of 41 noise-free latent frames ($48\times24\times42$) of the client's own video; one latent frame spans four video frames at 16\,fps, or sixteen engine ticks. Five convolutional blocks (three strided) map each latent frame from 48 to 512 channels, and a linear projection yields a token of width 1280. A causal transformer trunk (16 layers, 16 heads, pre-norm, $4\times$ feed-forward) processes these tokens, and a linear motion head on its outputs predicts the three-dimensional displacement. The sixteen control values per tick (thirteen buttons, pitch and yaw deltas, a validity flag) are projected to 256 dimensions, encoded by the action encoder (a four-layer causal transformer) and added to the trunk tokens. The place head takes detached trunk tokens, a pooled early feature map, and the unpooled $12\times21$ grid, concatenates a map embedding, and outputs an address (below). The trunk and action encoder have $338$M parameters and the motion and place heads $11$M.

\paragraph{Address.}
Positions are in the game engine's unit of length, u (not the token index $u$ of Appendix~\ref{app:field}); a player's body is 32\,u ($0.61$\,m) wide. To predict position as a class, each map is tiled into cells of $256\times256\times64$\,u, each divided $4\times4\times1$ into sub-cells; a cell, a sub-cell, and an offset of at most $\pm32$\,u from the sub-cell center represent a position exactly. The place head predicts the cell, then the sub-cell conditioned on the cell's 64-dimensional embedding, then the offset through a tanh. A map has at most $8{,}768$ sub-cells in its occupied cells.

\paragraph{Complementary filter.}
Eq.~(\ref{eq:position}), used for all results with predicted states, weights the integrated position and the place head's estimate equally at every latent frame and has no learned parameters. The estimate is initialized at the player's GT position at the start of the round (or of the window, for ten-second windows). The place head's estimates enter as displacements from its first estimate, so its constant bias cancels.

\paragraph{Within a block.}
As in networked games, each client extrapolates the motion of all players locally and is authoritative for its own player. The cameras of a block's four latent frames and the other players' positions over it are extrapolated from the controls~\citep{bernier2001latency}, since the block has no frames yet. Clients exchange player state once per block (\sect{sec:framework}); with predicted states, block $n$ starts from the positions published when it begins, each player's estimate at the last latent frame of block $n-1$. Each player's movement keys are integrated at 64\,Hz with per-key walking speeds, without collisions, and the facing direction accumulates the view controls from the last known facing direction. After the block, each client corrects only its own player's position, with the state model and Eq.~(\ref{eq:position}) on its new frames, and publishes it; every client then extrapolates over the next block from these corrected positions. With GT states, the GT cameras and positions are used.

\paragraph{Depth head.}
The depth head, a separate $44$M-parameter residual network with ten blocks at width 384 and a five-block half-resolution branch, predicts the depth-video latent of the client's own view from three latent frames of context (generated at inference), supervised offline by depth rendered from the public 3D map assets. A $0.21$M-parameter read-out maps this latent, without VAE decoding, to one depth value per $16\times16$ pixel patch, on which visibility (\sect{sec:grounding}) is computed; Fig.~\ref{fig:depth} shows the prediction decoded at full resolution on a recorded frame. The map assets contain no players, so the depth excludes occlusion between players. On held-out recorded frames of the clients of Table~\ref{tab:scene-agree}, the read-out depth differs from the rendered map depth by a median factor of $1.07$, with $85\%$ of patches within a factor of $1.25$. On generated frames, its effect is measured through retrieval (Table~\ref{tab:scene-intervals}).

\paragraph{Training.}
All components are trained on rounds disjoint from both evaluation sets, with AdamW, a one-cycle schedule, and weight decay $0.01$. The trunk and action encoder are trained with an $L_1$ displacement loss for 80{,}000 steps at batch 16 (about four hours of training) on $150$k windows from 8{,}317 rounds; the place head is then trained on the same windows, with the trunk frozen, for 60{,}000 steps at batch 16, both at a learning rate of $3\times10^{-4}$. The depth head is trained for 6{,}000 steps at a global batch of 192 on 32 GPUs, on $283$k windows from 10{,}469 rounds, with a learning rate of $4\times10^{-4}$ and an $L_1$ loss in latent space against the rendered map depth video, encoded by the same VAE. Per one-second block, the trunk with its heads takes $17$\,ms and the depth head with its read-out about $12$\,ms.

\paragraph{Contribution of each component.}
Table~\ref{tab:state} compares the state model with extrapolation from the controls, which networked games use between server updates~\citep{bernier2001latency}, and isolates its motion and place heads. The place head anchors the global position, height included. On recorded video the state model keeps the median error near $0.5$ body widths over the minute, with $87\%$ of clients within two body widths at $60$\,s, whereas extrapolation drifts to $7.6$ and leaves $2\%$; it does not know the height of the ground either, erring in height by a median of $3.4$ body widths over the minute against $0.06$ for the state model. On video generated with GT states, the state model's error, which includes each rollout's departure from the GT trajectory, grows to $2.6$ at $50$--$60$\,s, a third of that of extrapolation. The motion head reads local motion from the frames, such as a stop against a wall: in the Blocked windows it misplaces the one-second displacement by $0.05$ body widths on recorded and $0.38$ on generated video, against $2.38$ and $2.42$ for extrapolation, although alone it drifts, to $2.0$ and $22.7$ at $50$--$60$\,s. Eq.~(\ref{eq:position}) combines the two. Over the minute it is on par with the place head alone, which is slightly ahead on generated video ($2.51$ against $2.64$ body widths at $50$--$60$\,s); per latent frame, the step at which the complementary filter and the player state field are updated, its displacement error is about half that of the place head alone on recorded video, $0.21$ against $0.45$ body widths, and a third lower on generated video, $0.45$ against $0.69$.

\begin{table}[!htbp]
\caption{Position error in body widths (32\,u, $0.61$\,m) of the clients of the whole rounds of Table~\ref{tab:scene}, all alive at $60$\,s, on recorded video and on video generated by \method{} with GT states: per-client medians over each ten-second interval, then medians over clients, the share of clients within two body widths at $60$\,s, and, as Blocked, the median error of the one-second displacement over the one-second windows in which the controls imply more than one body width of motion and the GT position moves less than a third of it. Bold marks the better of the state model and extrapolation; the last two rows of each part isolate the motion and place heads of the state model. Extrapolation does not read the video; its two rows differ only in the instants at which the error is evaluated.}
\label{tab:state}
\centering
\footnotesize
\setlength{\tabcolsep}{5pt}
\begin{tabular*}{\linewidth}{@{\extracolsep{\fill}}lcccc@{}}
\toprule
& 0--10\,s $\downarrow$ & 50--60\,s $\downarrow$ & \shortstack[c]{Within two body\\widths at 60\,s $\uparrow$} & Blocked $\downarrow$ \\
\midrule
\multicolumn{5}{l}{\emph{Recorded video}} \\
State model, Eq.~(\ref{eq:position}) & \textbf{0.46} & \textbf{0.49} & \textbf{0.87} & \textbf{0.19} \\
Extrapolation from the controls & 1.86 & 7.63 & 0.02 & 2.38 \\
Place head only & 0.56 & 0.56 & 0.87 & 0.24 \\
Motion head only & 0.39 & 2.00 & 0.53 & 0.05 \\
\midrule
\multicolumn{5}{l}{\emph{Generated video, \method{}}} \\
State model, Eq.~(\ref{eq:position}) & \textbf{1.49} & \textbf{2.64} & \textbf{0.39} & \textbf{1.91} \\
Extrapolation from the controls & 1.86 & 7.60 & 0.03 & 2.42 \\
Place head only & 1.57 & 2.51 & 0.42 & 1.80 \\
Motion head only & 1.20 & 22.74 & 0.00 & 0.38 \\
\bottomrule
\end{tabular*}
\end{table}

\FloatBarrier
\section{Distributed deployment in detail}
\label{app:deploy}

This appendix specifies the client protocol, the measurements behind Fig.~\ref{fig:deployment} (Table~\ref{tab:distributed}), the attention patterns of the joint-generation methods, and the cost of a client over a long stream and of training a joint-generation method.

\paragraph{Deployment protocol.} Every client, once per block, publishes one message and reads the latest message of every other client, without a central process that merges them. For block $n$, client $j$ (i) publishes its player state $\mathbf{s}^j_n$, that is, the positions that Eq.~(\ref{eq:position}) estimates at the four latent frames of block $n-1$, the facing direction, and the controls and weapon for block $n$; with scene state, also block $n-1$ as a memory entry (four cameras and depths) and the memory entries it withdraws; and (ii) generates block $n$: it extrapolates the other players from their latest player states over the block (Appendix~\ref{app:state}), applies the withdrawals and retrieves at most one memory entry completed before block $n$, whose latents it fetches from the client that generated it. A message takes $3.7$\,kB for the player state and $8.3$\,kB for a memory entry (cameras in single precision, a $4\times24\times42$ depth in half precision); fetching the latents of a memory entry takes $387$\,kB. With GT states, player states come from the recording, and messages carry only the scene state. In our evaluations, each client runs on its own GPU, messages and latents are exchanged through a shared file system (for the timings of Table~\ref{tab:distributed} with $P\le8$, through shared memory on one machine), and the clients advance in lockstep, so that every client reads the messages of the same block.

\begin{table}[!htbp]
\caption{Serving $P$ players on H200 GPUs. \method{} runs one client per player, each on its own GPU, with scene state, as deployed with CUDA graphs, which reproduce its output bit for bit. All $P$ clients run concurrently and exchange messages and memory entries, through shared memory on one machine for $P\le8$ and through a shared file system across two machines for $P=10$ and $16$. A client's frame rate counts its own time per block and excludes the time it waits for the other clients at the lockstep of our evaluations, since, without lockstep, a client reads the latest messages (Appendix~\ref{app:deploy}); the table reports the median over clients and blocks. Gamma-World (2.2B parameters, $320\times480$; its distilled four-step causal student with a KV cache, the fastest configuration it releases) and Solaris (1.8B, $352\times640$) run as released, with random weights, all players on one GPU, which is their design. FPS per player includes decoding; memory per GPU is the peak of each framework's allocator, for \method{} the largest over clients. At $P=16$, a \method{} client receives 3.5\,Mb/s on average: per block, a 12.0\,kB message from each other client and, when it retrieves another client's memory entry, that entry's 387\,kB of latents. Traffic is received per GPU; for Gamma-World and Solaris it is that of splitting the players one per GPU. Attention compute is the attention FLOPs needed to generate one second of video at 16\,fps for all players on the GPU, computed analytically from the query--key products, $2d$ operations per pair, $d$ being the model width; $^*$the configuration runs out of memory. OOM: the configuration exceeds the memory of one H200.}
\label{tab:distributed}
\centering
\footnotesize
\setlength{\tabcolsep}{3.5pt}
\begin{tabular*}{\linewidth}{@{\extracolsep{\fill}}lccccc@{}}
\toprule
Players $P$ & 2 & 4 & 8 & 10 & 16 \\
\midrule
\multicolumn{6}{l}{\emph{FPS per player, including decoding} $\uparrow$} \\
\method{}, one GPU per player & 23.4 & 23.3 & 22.5 & 21.7 & 18.7 \\
Gamma-World, one GPU & 13.0 & 7.7 & 4.1 & 3.3 & 2.2 \\
Solaris, one GPU & 14.2 & 3.4 & 1.1 & 0.7 & OOM \\
\midrule
\multicolumn{6}{l}{\emph{Memory per GPU, GiB}} \\
\method{}, one GPU per player & 27.1 & 27.0 & 27.0 & 27.1 & 27.1 \\
Gamma-World, one GPU & 13.4 & 21.9 & 38.6 & 47.0 & 72.2 \\
Solaris, one GPU & 14.6 & 22.4 & 46.2 & 64.2 & OOM \\
\midrule
\multicolumn{6}{l}{\emph{Traffic received per GPU} $\downarrow$} \\
\method{}, one GPU per player & 2.2 Mb/s & 2.3 Mb/s & 2.7 Mb/s & 2.9 Mb/s & 3.5 Mb/s \\
Gamma-World, one GPU per player & 149 Mb/s & 227 Mb/s & 273 Mb/s & 285 Mb/s & 310 Mb/s \\
Solaris, one GPU per player & 26 Gb/s & 78 Gb/s & 182 Gb/s & 234 Gb/s & 389 Gb/s \\
\midrule
\multicolumn{6}{l}{\emph{Attention compute per GPU, TFLOP/s} $\downarrow$} \\
\method{}, one GPU per player & 5.9 & 5.9 & 5.9 & 5.9 & 5.9 \\
Gamma-World, one GPU & 40.7 & 81.4 & 163 & 203 & 326 \\
Solaris, one GPU & 34 & 137 & 548 & 856 & 2{,}192$^*$ \\
\bottomrule
\end{tabular*}
\end{table}

\paragraph{Joint-generation methods.} In \emph{Solaris}, each player attends to the windows of all players in one joint attention; \emph{frame concatenation} attends over the same tokens with the players' frames interleaved in time; Gamma-World's \emph{sparse hub} adds eight hub tokens per frame that attend to every player, while each player attends to its own window and the hubs. As released, Solaris and Gamma-World generate every player's view on one GPU, and we time them unchanged, all players on one GPU; frame concatenation is retrained only with our backbone (Appendix~\ref{app:scaling}) and is not timed. Joint attention reads the keys and values of all players at every layer, so, placed one player per GPU, they would exchange keys and values at every layer of every denoising step (Table~\ref{tab:distributed}). \method{} places each player on its own GPU.

\paragraph{Long streams.}
Because the attention window is bounded, a client's time per block and GPU memory footprint do not keep growing with the length of a stream. Over a three-minute stream of a single client with scene state and CUDA graphs, a block whose window holds memory frames takes 633\,ms in the first minute and 654 and 650\,ms in the second and third, the increase being mostly the writes to the memory bank once the pool of the client's own memory entries is full; the footprint grows once, by $0.9$\,GiB, when a CUDA graph is captured for the first window without memory frames, and not afterwards.

\paragraph{Training a joint-generation method.} The training cost of \method{} does not depend on $P$, since a sample is always the window of one player: under teacher forcing, 41 context and 41 noisy latent frames, or, when the sample holds memory frames, the shorter window of Appendix~\ref{app:scene}. A joint-generation method must place all players in one sample, $82P$ frames for the same model. On one H200, with frozen parameters and gradient checkpointing, a forward and backward pass over one sample takes 11.3\,GiB beyond the 10.8\,GiB of weights for 82 frames, 22.5\,GiB for 164, 64.7\,GiB for 328 and 100.9\,GiB for 410, and runs out of memory for 656. \method{} stays at the first value for any number of players, whereas Solaris and frame concatenation need 22.5\,GiB at two players and 100.9\,GiB at five and do not fit in the memory of one GPU at eight.

\FloatBarrier
\section{Training stages and scaling}
\label{app:scaling}

\begin{figure}[!htbp]
\centering
\includegraphics[trim=7.25bp 1.25bp 3bp 0bp,clip,width=0.52\linewidth]{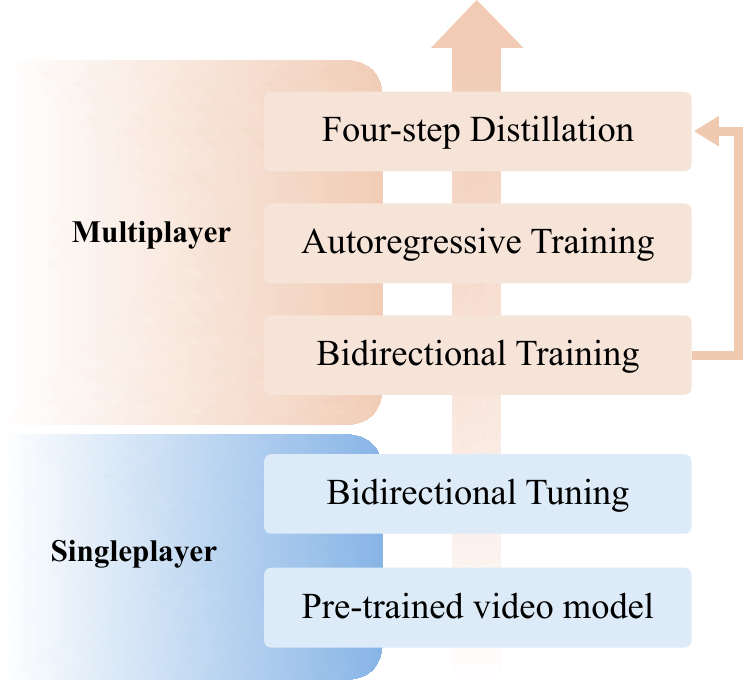}
\caption{Training stages, from a pretrained video model to four-step distillation.}
\label{fig:stages}
\end{figure}

This appendix details the training stages and runs (Fig.~\ref{fig:stages}, Table~\ref{tab:training}), the foreground weight, distillation, the compute and baselines of Table~\ref{tab:main}, and the scaling of the player state field with data (Table~\ref{tab:scaling}).

\begin{table}[!htbp]
\caption{Training runs of the reported models (Batch: global batch in windows); stages 3 and 4 are run once from each stage-2 model, with and without scene state. The ablations of Table~\ref{tab:ablations} retrain the stage-2 model without scene state from stage 1 for 6{,}000 steps at a batch of 64 (384k windows). \method{} denotes the stage-2 model without scene state at 25{,}000 steps in Table~\ref{tab:main}, its 6{,}000-step checkpoint in Table~\ref{tab:ablations} (the 384k column of Table~\ref{tab:scaling}a), and the four-step model with scene state in Tables~\ref{tab:scene}, \ref{tab:longhorizon}, \ref{tab:state}, \ref{tab:distributed}, \ref{tab:multiview}, \ref{tab:crowding} and~\ref{tab:scene-whole}--\ref{tab:scene-agree}.}
\label{tab:training}
\centering
\footnotesize
\setlength{\tabcolsep}{4pt}
\begin{tabular*}{\linewidth}{@{\extracolsep{\fill}}llrrr@{}}
\toprule
Stage & Initialization & Steps & Batch & Windows \\
\midrule
1: bidirectional tuning, five-second windows & Wan2.2-TI2V-5B & 34{,}000 & 256 & 8.70M \\
\quad then ten-second windows & five-second model & 6{,}000 & 256 & 1.54M \\
2: bidirectional training, player state field & stage 1 & 25{,}000 & 64 & 1.6M \\
\quad then scene state & stage 2 at 20{,}000 steps & 5{,}000 & 64 & 320k \\
3: autoregressive training & stage 2 & 5{,}000 & 64 & 320k \\
4: four-step distillation & stage 3 & 600 & 64 & 38.4k \\
\bottomrule
\end{tabular*}
\end{table}

\paragraph{Foreground weight.} From stage 2 on, the weight in Eq.~(\ref{eq:loss}) is $\alpha_k = 1 + (\lambda-1)\max_p g_p(k)\, \beta_p$ with $\lambda=3$, where $g_p$ is a unit-peak Gaussian centered on the visible other player $p$ on the $24\times42$ latent grid, and $\beta_p=\min\bigl(3,\max(1,\,1.5/r_p)\bigr)$ grows as the player's projected radius $r_p$ on that grid (at least half a cell) shrinks. The weight at a player's center is thus $3$ for a near player and at most $7$ for a distant one; the weight map is normalized by its batch mean.

\paragraph{Distillation.} Stage 4 runs 600 steps at a global batch of 64 ten-second windows, updating the generator every fifth step and the critic otherwise. The generator is initialized from the stage-3 model, and the teacher and critic from \method{}, the stage-2 model at 25{,}000 steps (Table~\ref{tab:main}), which has no scene state. From the recorded first frame, the generator produces the other 40 latent frames of each window as ten blocks of four, each conditioned on its own earlier outputs; the teacher and critic score the whole window bidirectionally. Distillation windows contain no memory frames, yet the four-step model retains the use of memory frames learned in stages 2 and 3 (Appendix~\ref{app:scene}): after 30\,s, \method{} exceeds the same weights with scene state off at inference by $1.23$\,dB PSNR $[0.91, 1.55]$ (Table~\ref{tab:scene-intervals}).

\paragraph{Compute.} The three models of Table~\ref{tab:main} used 2{,}637 GPU-hours for \method{} (25{,}000 steps, global batch of 64 windows, 64 GPUs); 3{,}146 for frame concatenation, which continues a 5{,}000-step run without player states (a further 640k windows and 1{,}224 GPU-hours); and 3{,}110 for Solaris (12{,}500 steps in total, first on 16 and then on 32 GPUs, at a global batch of 64 view pairs, i.e.\ 128 windows). Each model sees 1.6M windows with player states.

\paragraph{Baselines.} The two baselines keep jointly generated views consistent through attention between them; they receive the player states as tokens, in training and at inference, so that they know where the other players are, as the field does; the comparison is between joint generation and the player state field, not between ways of feeding states. Each player's state enters as one token per latent frame, with the token contents of the model trained without camera alignment (Appendix~\ref{app:field}) and the GT visibility label, folded per latent frame as for the field. The tokens join the cross-attention context, their mean over players is added to the frame's action modulation, and training uses the uniform loss.

Every model that receives the states as tokens, coupled or not, renders below $0.1$, against $0.763$ for \method{} at the same 384k windows: the model trained without camera alignment (Table~\ref{tab:ablations}), a single view given the states as tokens, renders $0.093$, against $0.074$ for frame concatenation and $0.056$ for Solaris at 1.6M windows (Table~\ref{tab:main}). We attribute the gap to where the state enters. The state injector writes each player's state explicitly, aligned with the camera, at the tokens where the player appears, so that the pretrained DiT can draw the player there; attention between views must learn this placement implicitly and, at these budgets, does not. The model trained without camera alignment differs from the baselines in three further ways: it has the foreground weight, lacks the visibility label and sees 384k windows. Nor does the budget explain the gap: the baselines stay below $0.09$ at every budget, including 256k and 512k on either side of 384k, whereas \method{} renders $0.62$ of the other players at 64k windows and improves with data (Table~\ref{tab:scaling}).

Scored only on the other player of the pair that the joint-generation methods generate together, \method{} renders $0.84$, against $0.03$ and $0.02$ for frame concatenation and Solaris, over the views with at least eight scored pairs. Although that player's own view seldom shows its body, the joint-generation methods generate that view and receive its state token, so they have at least as much information about its position as \method{}.

\begin{table}[!htbp]
\caption{Scaling with training data on the ten-second windows of Table~\ref{tab:main}, with the same entity criterion. \textbf{(a)} The player state field. \textbf{(b)} The render rate of \method{}, frame concatenation and Solaris at the budgets at which all three were trained and scored, each at its own checkpoint. Frame concatenation starts from a 5{,}000-step run (640k windows) without player states; its budgets count only windows with states.}
\label{tab:scaling}
\centering
\footnotesize
\setlength{\tabcolsep}{4pt}
\begin{tabular*}{\linewidth}{@{\extracolsep{\fill}}lcccccccc@{}}
\toprule
\textbf{(a)} Training windows & 64k & 128k & 256k & 384k & 512k & 768k & 1.0M & 1.6M \\
\midrule
Rendered $\uparrow$ & 0.617 & 0.667 & 0.738 & 0.763 & 0.759 & 0.774 & 0.784 & \textbf{0.817} \\
Unmatched $\downarrow$ & 0.154 & 0.102 & 0.075 & 0.076 & 0.060 & 0.047 & 0.044 & \textbf{0.039} \\
PSNR $\uparrow$ & 15.31 & 15.51 & 15.70 & 15.54 & 15.66 & 15.86 & 15.98 & \textbf{16.17} \\
LPIPS $\downarrow$ & 0.469 & 0.456 & 0.448 & 0.455 & 0.444 & 0.436 & 0.428 & \textbf{0.418} \\
FVD $\downarrow$ & 53.0 & 44.3 & 43.3 & 45.7 & 42.1 & 43.3 & 42.9 & \textbf{39.7} \\
\midrule
\textbf{(b)} Training windows & & 128k & 256k & & 512k & 768k & & 1.6M \\
\midrule
\method{} & & \textbf{0.667} & \textbf{0.738} & & \textbf{0.759} & \textbf{0.774} & & \textbf{0.817} \\
Frame concatenation & & 0.065 & 0.065 & & 0.081 & 0.067 & & 0.074 \\
Solaris & & 0.045 & 0.060 & & 0.041 & 0.036 & & 0.056 \\
\bottomrule
\end{tabular*}
\end{table}

\paragraph{Consistency between views.} Joint generation couples the views of a window to keep them consistent. Table~\ref{tab:view-agree} compares the three models of Table~\ref{tab:main} on this property with the measure of Table~\ref{tab:scene-agree}, applied to the two views of each ten-second window. Over the last five seconds, the depths of \method{}'s two views agree on $0.207$ of pixels, against $0.170$ for frame concatenation and $0.179$ for Solaris; the recorded video reaches $0.309$.

\begin{table}[!htbp]
\caption{Agreement between the depths of the two views of a ten-second window at the same instant, for the three models of Table~\ref{tab:main} with GT states: share of pixels whose reprojected depths agree within the larger of $24$\,u and $5\%$, averaged over the windows; chance places a view's depth from the most distant instant of its window at its current camera. Differences are paired over windows; their $95\%$ bootstrap intervals over windows (10{,}000 draws) all exclude zero.}
\label{tab:view-agree}
\centering
\footnotesize
\setlength{\tabcolsep}{5pt}
\begin{tabular*}{\linewidth}{@{\extracolsep{\fill}}lcc@{}}
\toprule
& First 5\,s & Last 5\,s \\
\midrule
\method{} & 0.257 & 0.207 \\
Frame concatenation & 0.232 & 0.170 \\
Solaris & 0.233 & 0.179 \\
Recorded video & 0.317 & 0.309 \\
Chance, depth from the most distant instant & 0.117 & 0.116 \\
\midrule
\multicolumn{3}{@{}l}{\emph{\method{} minus}} \\
Frame concatenation & $+0.025$ & $+0.037$ \\
Solaris & $+0.023$ & $+0.028$ \\
\bottomrule
\end{tabular*}
\end{table}

\FloatBarrier
\section{Evaluation protocols and long horizons in detail}
\label{app:longhorizon}

\paragraph{Data.} All video comes from OpenCS2~\citep{blanchon2026opencs2} (CC-BY-4.0), which replays Counter-Strike~2 demos published on HLTV and renders the ten players' first-person views at $1280\times720$ and 32\,fps, with every player's controls and state at each 64\,Hz engine tick. We use four maps (Dust2, Mirage, Nuke, Ancient): 569 match-maps from 55 events with 11{,}853 rounds, split by match into 303 training matches (10{,}480 rounds) and 40 held-out matches (1{,}373 rounds). Models read the video at $384\times672$ and 16\,fps. The detector, YOLO11x, is fine-tuned on Dust2 frames labeled automatically with the pretrained detector's person boxes that match boxes projected from the GT positions. On Dust2 frames from matches unused in fine-tuning, labeled the same way, it reaches $0.87$ precision and $0.90$ recall at its F1-optimal threshold; evaluation uses a confidence of $0.05$.

\paragraph{Evaluation sets.} Both sets are drawn from the 1{,}373 rounds of the 40 held-out matches by rules fixed before any reported result and based only on the recordings and the map assets: the visibility labels computed from them, which players' first-person views the dataset holds, and the length of each recording. The set of ten-second windows takes 64 rounds per map, 256 in total, in a seeded random order and, in each, the ten-second window, among windows starting every 2.5\,s, and the pair of players with a first-person view of it that maximize the smaller of the two shares of frames in which one player has the other in view. These two players give the two generated views, the pair that the joint-generation methods generate together; this smaller share ranges from 16\% to 100\% (median 48\%), and windows start a median of 23\,s into the round. The whole-round set scores each group of three players whose recordings all last at least 60\,s by the share of seconds in which one player has another in view in at least half of the frames, averaged over its two pairs with the highest shares and multiplied by the rollout length capped at 120\,s; it keeps the eight rounds per map with the best-scoring groups. Although the rule does not require one team, in all 32 rounds the three selected players are teammates who remain alive throughout their rollouts, which run from the round start to the end of the shortest recording (67 to 120\,s). All methods and ablations use the same windows and rounds.

Tables~\ref{tab:longhorizon-all}, \ref{tab:state} and~\ref{tab:multiview} report the first 60\,s, which every client of these rounds reaches. Tables~\ref{tab:scene}, \ref{tab:crowding} and~\ref{tab:scene-whole} include every window to the end of each round.

\paragraph{Whole rounds (Tables~\ref{tab:scene} and~\ref{tab:longhorizon}a).} The whole-round results of Tables~\ref{tab:multiview} and~\ref{tab:crowding} are obtained with \method{}, the four-step model with scene state, and the whole-round rows of Table~\ref{tab:longhorizon-all} and the clips of Table~\ref{tab:vbench-sensitivity} with the four-step model trained without scene state, all on the 32 rounds of Table~\ref{tab:scene}. In every round, the three selected teammates are clients, sharing one memory bank when the model has scene state, and the other seven players follow their GT states; every client is rolled out from the round start, given the GT states of all players. In Table~\ref{tab:longhorizon}a, \method{} runs on the same rounds in closed-loop deployment, with all five players of the team as clients until their recordings end; in closed-loop deployment, every client runs the whole deployed loop with predicted states and takes nothing from the recording beyond its first frames and position: after every block, each client estimates its own position from its new frames with the state model and Eq.~(\ref{eq:position}) and publishes it, the other clients render that client's player at this position in their next block, and all visibility is predicted as in \sect{sec:grounding}.

\paragraph{Ten-second windows (Tables~\ref{tab:main} and~\ref{tab:predicted}).} Table~\ref{tab:predicted} evaluates the state model: its two rows use the same generator and differ only in whether the states are GT or predicted. Its generated views (one per client) differ from those of Table~\ref{tab:main} (two per window), so rates are compared only within a table. In both rows of Table~\ref{tab:predicted}, every player whose first-person view of the window is in the dataset runs a client; the others, nearly all dead before or during the window, keep their GT states. For the \emph{Predicted positions} row of Table~\ref{tab:predicted}, every client runs in closed loop: after every block, it estimates its own position from its new frames with the state model and Eq.~(\ref{eq:position}), initialized at the GT position, and publishes it; the other clients of the window render that client's player there in their next block, advancing it over the block by extrapolation from its controls (Appendix~\ref{app:state}), and all visibility is predicted as in \sect{sec:grounding} (Figs.~\ref{fig:window-pred} and~\ref{fig:window-pred-b}). Table~\ref{tab:main-ci} reports the corresponding paired differences with $95\%$ bootstrap intervals, all of which exclude zero, and Table~\ref{tab:main-ci-criterion} repeats them under other settings of the entity criterion.

\paragraph{One-hour rollouts.} Each of the 16 one-hour rollouts (four per map, for different players) runs one \method{} client with scene state, which estimates its own position from its generated video, on one player's recorded controls, with all other players following their GT states. It starts from the recorded first frame of the first round and continues over 66 to 102 consecutive rounds from four to six matches held out from training; each round lasts as long as the player's recording of it and continues from the last generated frame of the previous one. At the start of every round, the player respawns: the scene state is cleared and the position estimate restarts from the recorded spawn position. The rollout is cut at one hour.

\begin{table}[!htbp]
\caption{Paired differences on the ten-second windows of Tables~\ref{tab:main} and~\ref{tab:predicted}, whose 95\% intervals from 10{,}000 bootstrap resamples of the windows, with the views of a window resampled together, all exclude zero. PSNR, SSIM and LPIPS report mean per-view differences, while Rendered and Unmatched report median per-view differences, which need not equal the difference between the medians of these tables. FVD compares two distributions of clips and therefore has no paired difference.}
\label{tab:main-ci}
\centering
\footnotesize
\setlength{\tabcolsep}{5pt}
\begin{tabular*}{\linewidth}{@{\extracolsep{\fill}}lccccc@{}}
\toprule
& $\Delta$PSNR & $\Delta$SSIM & $\Delta$LPIPS & $\Delta$Rendered & $\Delta$Unmatched \\
\midrule
\method{} minus frame concatenation & $+0.80$ & $+0.015$ & $-0.042$ & $+0.631$ & $-0.571$ \\
\method{} minus Solaris & $+0.51$ & $+0.006$ & $-0.032$ & $+0.636$ & $-0.592$ \\
Predicted minus GT positions & $-0.38$ & $-0.007$ & $+0.021$ & $-0.058$ & $+0.032$ \\
\bottomrule
\end{tabular*}
\end{table}

\begin{table}[!htbp]
\caption{Table~\ref{tab:main-ci} under other settings of the entity criterion (Appendix~\ref{app:longhorizon}), each changed in the generated and the recorded video alike: \method{} minus frame concatenation (Concat.) and minus Solaris, and predicted minus GT positions (Pred.). Every difference keeps its sign, and its $95\%$ bootstrap interval, computed as in Table~\ref{tab:main-ci}, excludes zero.}
\label{tab:main-ci-criterion}
\centering
\footnotesize
\setlength{\tabcolsep}{5pt}
\begin{tabular*}{\linewidth}{@{\extracolsep{\fill}}lcccccc@{}}
\toprule
& \multicolumn{3}{c}{$\Delta$Rendered} & \multicolumn{3}{c}{$\Delta$Unmatched} \\
\cmidrule(lr){2-4}\cmidrule(lr){5-7}
Setting of the criterion & Concat. & Solaris & Pred. & Concat. & Solaris & Pred. \\
\midrule
Match within half a body width & $+0.648$ & $+0.656$ & $-0.095$ & $-0.692$ & $-0.709$ & $+0.101$ \\
Match within two body widths & $+0.553$ & $+0.547$ & $-0.038$ & $-0.364$ & $-0.409$ & $+0.012$ \\
Detector confidence $0.25$ & $+0.585$ & $+0.584$ & $-0.064$ & $-0.603$ & $-0.636$ & $+0.029$ \\
Detector confidence $0.5$ & $+0.556$ & $+0.552$ & $-0.090$ & $-0.616$ & $-0.624$ & $+0.025$ \\
Boxes of at least $48^2$ pixels & $+0.638$ & $+0.632$ & $-0.053$ & $-0.515$ & $-0.521$ & $+0.062$ \\
Mean instead of median & $+0.596$ & $+0.593$ & $-0.070$ & $-0.546$ & $-0.568$ & $+0.075$ \\
\bottomrule
\end{tabular*}
\end{table}

\paragraph{VBench.} Table~\ref{tab:longhorizon-all} reports every VBench dimension computed from video alone; Table~\ref{tab:vbench-sensitivity} measures, on 24 random ten-second windows of the whole rounds, which of them, except temporal flickering, respond to image damage. Under seven corruptions, imaging and aesthetic quality fall by up to $48\%$ and $20\%$, whereas subject consistency, background consistency and motion smoothness change by at most $3.5\%$ and rise when frames are held, and dynamic degree responds only to held frames.

\begin{table}[!htbp]
\caption{Reference-free VBench scores of the ten-second windows of the whole rounds (GT states, four-step model trained without scene state) and of the one-hour rollouts of \method{} with scene state, per twenty-minute interval. Dynamic degree is the fraction of clips classified as moving.}
\label{tab:longhorizon-all}
\centering
\footnotesize
\setlength{\tabcolsep}{4pt}
\begin{tabular*}{\linewidth}{@{\extracolsep{\fill}}lccccccc@{}}
\toprule
& Subject & Background & Motion & Flicker & Dynamic & Imaging & Aesthetic \\
\midrule
\multicolumn{8}{l}{\emph{Whole rounds}} \\
0--10\,s & 0.732 & 0.888 & 0.953 & 0.926 & 1.000 & 70.8 & 0.480 \\
10--20\,s & 0.792 & 0.917 & 0.965 & 0.947 & 1.000 & 71.1 & 0.491 \\
20--30\,s & 0.822 & 0.929 & 0.966 & 0.948 & 0.969 & 71.5 & 0.499 \\
30--40\,s & 0.824 & 0.926 & 0.966 & 0.948 & 0.948 & 71.8 & 0.497 \\
40--50\,s & 0.827 & 0.929 & 0.964 & 0.946 & 0.958 & 72.1 & 0.497 \\
50--60\,s & 0.829 & 0.926 & 0.966 & 0.948 & 0.979 & 72.0 & 0.495 \\
\midrule
\multicolumn{8}{l}{\emph{One-hour rollouts}, recorded controls} \\
0--20\,min & 0.803 & 0.910 & 0.965 & 0.947 & 0.950 & 70.3 & 0.486 \\
20--40\,min & 0.797 & 0.912 & 0.965 & 0.946 & 0.975 & 70.3 & 0.489 \\
40--60\,min & 0.797 & 0.910 & 0.965 & 0.947 & 0.966 & 69.2 & 0.486 \\
\bottomrule
\end{tabular*}
\end{table}

\begin{table}[!htbp]
\caption{VBench sensitivity: percent change under paired corruptions of random ten-second windows of the whole rounds, generated by the four-step model trained without scene state (Table~\ref{tab:longhorizon-all}).}
\label{tab:vbench-sensitivity}
\centering
\footnotesize
\setlength{\tabcolsep}{4pt}
\begin{tabular*}{\linewidth}{@{\extracolsep{\fill}}lcccccc@{}}
\toprule
Damage & Subject & Background & Motion & Dynamic & Aesthetic & Imaging \\
\midrule
Blur, $\sigma=0.5$\,px & $-0.1$ & $+0.6$ & $+0.3$ & $0.0$ & $-0.1$ & $-5.7$ \\
Blur, $\sigma=2$\,px & $-1.1$ & $+1.3$ & $+1.1$ & $0.0$ & $-15.7$ & $-48.1$ \\
JPEG, quality 30 & $-0.5$ & $-2.1$ & $-0.2$ & $0.0$ & $-7.7$ & $-6.3$ \\
JPEG, quality 12 & $-1.6$ & $-0.7$ & $-0.3$ & $0.0$ & $-20.4$ & $-19.6$ \\
Noise, $\sigma=10$ & $-0.8$ & $-1.4$ & $-0.8$ & $0.0$ & $-8.6$ & $-14.9$ \\
Gamma $1.5$ & $-0.1$ & $-0.1$ & $+0.1$ & $0.0$ & $-1.3$ & $-7.4$ \\
Every 8th frame held & $+3.5$ & $+2.3$ & $+2.0$ & $-37.5$ & $+0.8$ & $+0.8$ \\
\bottomrule
\end{tabular*}
\end{table}

\paragraph{The entity criterion.}
Rendered and Unmatched are computed in all tables with the rule of \sect{sec:criterion}, against the state given to the generator (with GT states, the recorded position and the visibility computed from the map assets). A pair of another player and a frame is scored if the player is marked visible in the state given to the generator, the player's box clipped to the $384\times672$ frame covers at least $24^2$ pixels, and the detector finds the player in the recorded video; the player is rendered if a person is detected within one body width, the width of the player's projected box, of its position in the given player state, each detected person matching at most one player. A detected person is unmatched if neither a player marked visible in the given state, with a projected box at least 24 pixels on its short side, nor a person detected in the recorded video lies within one body width of it; the unmatched rate is the unmatched share of persons detected in the generated frames. Both rates are computed for each view, one player's view of a ten-second window, with a whole round split into consecutive ten-second windows, and are summarized by their median over the views with at least eight pairs or eight detected persons, respectively; the views in the unmatched median thus depend on how many persons a generator draws and can differ between rows. Table~\ref{tab:crowding} instead pools the scored pairs of each level of crowding.

\paragraph{Agreement on player positions.}
To evaluate whether independently generated views place the same player at the same location, Table~\ref{tab:multiview} measures, for two clients of a round, whether both render the player of the third client and where. The evaluation includes the frames in which that player is marked visible for both clients, its clipped box covers at least $24^2$ pixels in both views, and the detector finds it in both recordings. A client is considered to render that player if a person is detected within one body width of the player's GT position.

\begin{table}[!htbp]
\caption{Agreement of two clients on the position of the player of the third client of their round, for \method{} on the whole rounds with GT states. Frames denotes the number of frames in which that player is scored in both views, and Rendered by both the fraction of these frames in which both clients render it. Disagreement reports, over the frames rendered by both, the median closest distance between the two clients' viewing rays to the persons matched to that player (one body width is 32\,u).}
\label{tab:multiview}
\centering
\footnotesize
\setlength{\tabcolsep}{6pt}
\begin{tabular*}{\linewidth}{@{\extracolsep{\fill}}lccc@{}}
\toprule
& Frames & Rendered by both & Disagreement (u) $\downarrow$ \\
\midrule
0--10\,s & 955 & 0.687 & 27.4 \\
10--20\,s & 492 & 0.626 & 18.7 \\
20--30\,s & 652 & 0.644 & 19.3 \\
30--40\,s & 831 & 0.712 & 17.0 \\
40--50\,s & 1{,}033 & 0.804 & 19.2 \\
50--60\,s & 642 & 0.715 & 16.7 \\
\bottomrule
\end{tabular*}
\end{table}

\paragraph{Crowded frames.}
To examine whether rendering degrades in crowded views, Table~\ref{tab:crowding} groups the whole-round render rate of \method{} by the number of other players in view, counting the players marked visible whose clipped box covers at least $24^2$ pixels. The render rate remains between $0.79$ and $0.82$ for one to four other players, and no frame of the 32 rounds contains more.

\begin{table}[!htbp]
\caption{\method{} on the whole rounds by the number of other players in view. Rendered is pooled over the scored pairs of each level of crowding, since one window mixes levels of crowding.}
\label{tab:crowding}
\centering
\footnotesize
\setlength{\tabcolsep}{6pt}
\begin{tabular*}{\linewidth}{@{\extracolsep{\fill}}lcccc@{}}
\toprule
Other players in view & 1 & 2 & 3 & 4 \\
\midrule
Frames & 44{,}395 & 15{,}274 & 3{,}498 & 631 \\
Scored pairs & 31{,}525 & 21{,}980 & 7{,}540 & 1{,}941 \\
Rendered $\uparrow$ & 0.809 & 0.797 & 0.795 & 0.821 \\
\bottomrule
\end{tabular*}
\end{table}
\FloatBarrier
\section{Scene state in detail}
\label{app:scene}

The 32 rounds of Table~\ref{tab:scene} come from held-out matches on the training maps, as the place head requires, but outside the training data of the state model and of every generator stage. All rows of Table~\ref{tab:scene} use the same ten-second windows of the whole rounds.

Each ablation changes one aspect of \method{} on the same rounds. \emph{Scene state off at inference} keeps \method{}'s weights and removes only the memory frames, isolating the scene state; the \emph{model trained without scene state} also has clean context, as in its training (\emph{Training} below). After 30\,s, \method{} exceeds both on PSNR, SSIM and LPIPS (Table~\ref{tab:scene-intervals}; examples in Figs.~\ref{fig:scene-mirage} and~\ref{fig:scene-dust2}); in the first 30\,s, the model trained without scene state is slightly ahead, with intervals that include zero. The other ablations change what is retrieved: \emph{Only its own earlier blocks} restricts retrieval to the client's own memory entries, \emph{Least-covering memory} selects the memory entry covering the fewest missing pixels, and \emph{GT frames in the memory bank} replaces the other clients' memory entries with their GT frames, which also supply the depth that locates them, keeping which memory entries exist and when.

\begin{table}[!htbp]
\caption{Table~\ref{tab:scene} over whole rounds, against the recorded video. PSNR, SSIM and LPIPS: means over the ten-second windows; FVD: over their clips; Rendered and Unmatched: medians over views (Appendix~\ref{app:longhorizon}).}
\label{tab:scene-whole}
\centering
\footnotesize
\setlength{\tabcolsep}{4pt}
\begin{tabular*}{\linewidth}{@{\extracolsep{\fill}}lcccccc@{}}
\toprule
& FVD $\downarrow$ & PSNR $\uparrow$ & SSIM $\uparrow$ & LPIPS $\downarrow$ & Rendered $\uparrow$ & Unmatched $\downarrow$ \\
\midrule
Trained without scene state & 78.4 & 13.60 & 0.403 & 0.572 & 0.815 & 0.053 \\
Scene state off at inference & 90.5 & 13.40 & 0.402 & 0.593 & 0.820 & 0.049 \\
Only its own earlier blocks & 91.1 & 13.87 & 0.407 & 0.574 & 0.825 & 0.042 \\
Least-covering memory & 98.5 & 13.21 & 0.393 & 0.602 & 0.804 & 0.054 \\
\method{} & 89.3 & 14.40 & 0.416 & 0.545 & 0.841 & 0.037 \\
\midrule
GT frames in the memory bank & 75.0 & 14.99 & 0.433 & 0.513 & 0.853 & 0.030 \\
\bottomrule
\end{tabular*}
\end{table}

\begin{table}[!htbp]
\caption{\method{} minus each other row of Table~\ref{tab:scene}, paired over the same ten-second windows: mean differences; $^\dagger$ marks a $95\%$ cluster-bootstrap interval over rounds (10{,}000 draws) that includes zero. Positive $\Delta$PSNR and $\Delta$SSIM and negative $\Delta$LPIPS favor \method{}. The last row, \emph{GT frames in the memory bank}, whose memory entries hold GT frames that clients cannot have, is ahead (negative $\Delta$PSNR and $\Delta$SSIM, positive $\Delta$LPIPS).}
\label{tab:scene-intervals}
\centering
\footnotesize
\setlength{\tabcolsep}{4pt}
\begin{tabular*}{\linewidth}{@{\extracolsep{\fill}}lccc@{}}
\toprule
\method{} minus & $\Delta$PSNR & $\Delta$SSIM & $\Delta$LPIPS \\
\midrule
\multicolumn{4}{@{}l}{\emph{First 30\,s}} \\
Trained without scene state & $-0.05^\dagger$ & $-0.0078^\dagger$ & $+0.0138^\dagger$ \\
Scene state off at inference & $+0.51$ & $+0.0072^\dagger$ & $-0.0248$ \\
Only its own earlier blocks & $+0.32$ & $+0.0059^\dagger$ & $-0.0189$ \\
Least-covering memory & $+0.61$ & $+0.0097$ & $-0.0293$ \\
GT frames in the memory bank & $-0.29$ & $-0.0113$ & $+0.0163$ \\
\midrule
\multicolumn{4}{@{}l}{\emph{After 30\,s}} \\
Trained without scene state & $+1.20$ & $+0.0227$ & $-0.0466$ \\
Scene state off at inference & $+1.23$ & $+0.0178$ & $-0.0588$ \\
Only its own earlier blocks & $+0.63$ & $+0.0102$ & $-0.0347$ \\
Least-covering memory & $+1.46$ & $+0.0294$ & $-0.0712$ \\
GT frames in the memory bank & $-0.75$ & $-0.0188$ & $+0.0397$ \\
\bottomrule
\end{tabular*}
\end{table}

\paragraph{Agreement on the scene.}
Since each rollout departs from the recording, the scene state is also evaluated by the consistency among clients within the world they generate. Mat.\ Pix.\ counts the pixels that GIM~\citep{shen2024gim} matches between two clients' frames at the same instant with confidence above 0.05, as in SynCamMaster~\citep{bai2025syncammaster}. Inliers counts the LoFTR~\citep{sun2021loftr} matches between the two frames that fit one essential matrix, estimated with MAGSAC++~\citep{barath2020magsac}, and RotErr, as in CameraCtrl~\citep{he2025cameractrl} and SynCamMaster, is the angle between the rotation recovered from that matrix and the relative rotation of the two cameras. All three are computed over the pairs of clients whose views are co-visible in the recording. On the recorded video of the pairs scored in Table~\ref{tab:longhorizon}a, same-instant pairs score far better than pairs offset by 10\,s (Mat.\ Pix.\ $163.1$K against $52.3$K, Inliers $128.3$ against $36.4$, RotErr $16.7^\circ$ against $64.9^\circ$; better in all but two rounds for Mat.\ Pix.\ and in every round for the others), so the measures track agreement between views. To test whether the clients of a round generate the same scene, Table~\ref{tab:scene-agree} compares the depths that two clients predict for their generated frames at the same instant. The depth that the depth head predicts for one client's latent frame is back-projected with that client's camera and reprojected into the other client's camera; a pixel agrees if the reprojected depth and the other client's own depth there lie within the larger of $24$\,u and $5\%$. The share of agreeing pixels is averaged over the same pairs and over the rounds. The recorded video gives the upper bound, and a client's depth from 10\,s later, placed at its current camera, the chance level. The scene state raises this agreement, particularly after 30\,s, and every ablation falls below \method{}. It leaves each client's agreement with its own next latent frame unchanged, so the gain does not come from a steadier depth estimate. In the closed-loop deployment of Table~\ref{tab:longhorizon}a, where the clients use predicted states and each client's depth is reprojected with the cameras it publishes, the scene state likewise raises the agreement after 30\,s, from $0.092$ to $0.120$, and leaves it unchanged in the first 30\,s. On the same pairs, relative to scene state off, \method{} changes Mat.\ Pix.\ by $+20.1$K $[5.1, 35.7]$ and $+13.2$K $[4.5, 22.5]$, Inliers by $+5.2$ $[1.1, 9.3]$ and $+9.0$ $[5.1, 13.1]$, and RotErr by $-2.8^\circ$ $[-4.8, -0.9]$ and $-3.4^\circ$ $[-6.7, -0.3]$ in the first 30\,s and after 30\,s, respectively, with $95\%$ round-cluster bootstrap intervals. Every view shows the weapon and hands at the same screen position; with the lower right of each frame and its lower-left corner masked, the differences after 30\,s remain $+8.0$K $[2.2, 14.1]$ in Mat.\ Pix., $+5.2$ $[2.7, 7.9]$ in Inliers and $-12.9^\circ$ $[-22.1, -4.8]$ in RotErr, and those in the first 30\,s also exclude zero.

\begin{table}[!htbp]
\caption{Agreement between the depths predicted for two clients' generated frames at the same instant, on the whole rounds of Table~\ref{tab:scene} with GT states and, in the lower part, in the closed-loop deployment of Table~\ref{tab:longhorizon}a with the cameras each client publishes: share of pixels whose reprojected depths agree within the larger of $24$\,u and $5\%$; chance places a client's depth from 10\,s later at its current camera. Differences are paired over rounds; $^\dagger$ marks a $95\%$ cluster-bootstrap interval over rounds (10{,}000 draws) that includes zero. Own next latent frame applies the same measure to each client and its own next latent frame; scene state does not change it, so its gain between clients is not a steadier depth estimate.}
\label{tab:scene-agree}
\centering
\footnotesize
\setlength{\tabcolsep}{5pt}
\begin{tabular*}{\linewidth}{@{\extracolsep{\fill}}lcc@{}}
\toprule
& First 30\,s & After 30\,s \\
\midrule
\multicolumn{3}{@{}l}{\emph{GT states}} \\
Trained without scene state & 0.259 & 0.135 \\
Scene state off at inference & 0.229 & 0.133 \\
Only its own earlier blocks & 0.231 & 0.147 \\
Least-covering memory & 0.225 & 0.119 \\
\method{} & 0.264 & 0.206 \\
Recorded video & 0.411 & 0.365 \\
Chance, depth from 10\,s later & 0.113 & 0.101 \\
\midrule
\multicolumn{3}{@{}l}{\emph{\method{} minus scene state off}} \\
Between clients & $+0.035$ & $+0.073$ \\
Own next latent frame & $+0.004^\dagger$ & $+0.000^\dagger$ \\
\midrule
\multicolumn{3}{@{}l}{\emph{Closed-loop deployment, predicted states}} \\
\method{} & 0.247 & 0.120 \\
Scene state off at inference & 0.243 & 0.092 \\
\multicolumn{3}{@{}l}{\emph{\method{} minus scene state off}} \\
Between clients & $+0.004^\dagger$ & $+0.028$ \\
Own next latent frame & $-0.001^\dagger$ & $-0.002^\dagger$ \\
\bottomrule
\end{tabular*}
\end{table}

\paragraph{FVD.} FVD compares the generated and recorded ten-second windows of the whole rounds as unpaired distributions. It is similar for \method{} and \emph{Scene state off at inference} (89.3 and 90.5), which share weights; the lower FVD of the model trained without scene state (78.4) thus stems from its weights or its clean context.

\paragraph{Ray embedding.}
The MLP $\phi$ of \sect{sec:scene} is Linear--SiLU--Linear with a zero-initialized last layer, so the ray embedding starts at zero; its $9.5$M parameters are trained with the full backbone. The length scale is $\ell = 420$\,u, and each viewing direction uses its frame's field of view.

\paragraph{Managing the memory entries.} All clients share one scene state per round. A memory entry is one block (four latent frames, one second of video), stored with each latent frame's camera and the depth predicted by the depth head from the block's own latents, one value per $16\times16$-pixel patch (\sect{sec:grounding}). Each latent frame spans four video frames and takes the camera and depth of the last, when its content is complete and the state model reads its position. A set of memory entries covers a pixel of a frame if the nearest of their reprojected points matches the frame's depth there within the larger of $24$\,u and $5\%$. Every generated block is published as a memory entry; each client keeps at most $B=64$ of its own, withdrawing beyond this bound the one best covered by its others (the older on a tie). With reprojections between a client's memory entries cached, publishing adds at most $2B$ reprojections and withdrawal requires none. The scene state thus holds at most $B$ memory entries per client for any rollout length, and $BP$ for $P$ clients. Retrieval projects every memory entry completed before a block, so its per-block cost grows with the number of memory entries, at most linearly in $P$, unlike the generator's; the frame rates of Table~\ref{tab:distributed} include it.

\paragraph{Retrieval.} Retrieval runs before the first denoising step, when the block has no frames, and so uses only its four cameras and its missing pixels, which no 3D point of the client's twelve most recent latent frames reaches. A memory entry covers a missing pixel if its point lies, within the tolerance, on the visible surface given by the nearest point of any memory entry. Of all memory entries completed before the block, the client's own included, the one covering the most missing pixels is retrieved (the more recent on a tie), and none if no memory entry covers any. Retrieving at most one memory entry bounds the generator's input at twenty latent frames (four memory frames, twelve recent context frames, four target frames) for any number of clients. In the rollouts of Table~\ref{tab:scene}, a memory entry is retrieved for $95\%$ of the blocks that read the scene state, $0.65$ $[0.61, 0.68]$ of them from the other two clients; in the closed-loop deployment of Table~\ref{tab:longhorizon}a, for $96\%$, $0.47$ $[0.44, 0.50]$ of them from the other clients. Replayed on the generated blocks of the whole rounds whose view, by the map geometry, shows surface unseen from the recent context, with coverage measured on the map geometry rather than on the predicted depth that retrieval uses, the retrieved memory entry attains $0.81$ $[0.78, 0.83]$ of the coverage of the best earlier memory entry of the other clients, against $0.70$ $[0.67, 0.72]$ for the memory entry outside the recent context whose camera is nearest the block's; the paired difference, $+0.11$ $[+0.09, +0.13]$, grows with head rotation. In every evaluation the scene state is read from the 25th latent frame of a rollout onward, so the first six seconds use the recent context alone.

\paragraph{Training.}
With probability $0.8$, a training sample includes GT frames as memory frames: the four latent frames of a same-round teammate's view that end no later than the last target frame and show most of what the recent context lacks, as counted by $m_k$; otherwise it has none, so that the generator also learns to generate without scene state. In stage 3, under block-causal attention, the memory frames precede the twelve most recent and four target latent frames; in stage 2, they precede 32 recent latent frames, noised with the target, and four target latent frames, all under bidirectional attention. The loss at position $k$ of target frame $f$ is multiplied by $\alpha_k c_k$ in Eq.~(\ref{eq:loss}), with $c_k = \tfrac12\bigl(1 + m_k/\bar m_f\bigr)$. The mask $m_k \in \{0,1\}$ marks the positions whose surface the memory frames show and the recent context does not, computed during training from the map geometry and the recorded cameras, and $\bar m_f$ is the share of marked positions in frame $f$; in a frame without marked positions, $c_k = 1$. Otherwise, marked positions receive the weight $\tfrac12(1 + 1/\bar m_f)$ and the other positions $\tfrac12$, so that $c_k$ averages to one over each frame and the total weight of the frame is unchanged. The recent context frames of the model with scene state are noised to a timestep drawn between 16 and 32 of 1{,}000 in stage 3, and to 16 in distillation and at inference.
\FloatBarrier
\section{The player state field in detail}
\label{app:field}

This appendix specifies the player state field of \sect{sec:field}.

\paragraph{Channels.}
The field has twenty-three channels on the model's token grid, $12\times21$ per latent frame (the $24\times42$ latent grid of a $384\times672$ frame, patchified by two): coverage; log-compressed depth; the sine and cosine of the yaw relative to the observer; team; headcount; nine control channels, each the fraction of the latent frame during which a key is held (forward, back, strafe left, strafe right, jump, crouch, walk, attack, reload); four weapon-embedding channels; dying and corpse channels, written where the player fell; and two identity channels.

\paragraph{Splatting.}
Each player to be written is splatted onto the token grid with the isotropic Gaussian kernel of~\citet{wang2025ati}, centered on the body at the projected position raised by the projected radius (the engine's origin is the ground between the feet). The weight of player $p$ at token $u$ is $w_p(u) = e_p \exp\bigl(-\lVert \mathbf{x}_u - \boldsymbol{\mu}_p \rVert^2/2\sigma^2\bigr)$, with $\sigma$ half a token for all players; $e_p$ is zero for a player not written and otherwise a confidence between $0.3$ and $1$ that smooths visibility over the latent frames of a block. Player size enters through the depth channel, not the kernel. The spread of the weights over neighboring tokens encodes the sub-token position of the center: with the coarse field of Table~\ref{tab:ablations}, which resolves position only to $2\times2$ tokens, the render rate is $0.708$, against $0.763$ for \method{}.

\paragraph{Composition.}
At each token the field keeps the two players with the largest depth-discounted weight $w_p(u)\,e^{-\tilde z_p}$, with $\tilde z_p$ the log-compressed depth, and combines them as
\begin{equation}
c(u) = \min\Bigl(1,\; \sum_{p\in S(u)} w_p(u)\Bigr),
\qquad
\mathbf{F}_a(u) = \sum_{p\in S(u)} w_p(u)\, a(s_p),
\end{equation}
where $S(u)$ is the selected pair, $c$ the coverage channel and $a(s)$ the value of a state in channel $a$. The headcount channel sums each player's peak weight in the frame, capped at four. A frame without players to write is exactly zero in all channels.

\paragraph{Visibility.}
Another player is written into the field only if it lies in the view frustum and is visible (\sect{sec:grounding}). Visibility is tested at the feet raised by $40$\,u and at eight points on a horizontal circle of radius $40$\,u around that point: the player is visible if, at any test point, the 3D points of the recent context frames reach its projection and their median depth over the $2\times2$ depth-map cells there (each $16\times16$ pixels) lies at most $24$\,u in front of it. Where no 3D point reaches, the player is not written at the first denoising step; after it, visibility is re-tested once on the depth of the block's clean estimate $\hat{\mathbf{x}}_0$ and held for the remaining steps. In evaluations using GT states, the GT visibility labels, computed from the recorded positions and the public 3D map assets, are used instead.

\paragraph{Injection.}
The field enters after the second of the 30 DiT blocks through one zero-initialized stem of thirty-two channels, so stage 2 starts from the stage-1 generator; the stem's learning rate is $1.4\times10^{-3}$, the backbone's $2.8\times10^{-5}$. The remaining 28 DiT blocks process the field; injected after the 23rd DiT block, it passes through only seven, and the render rate falls from $0.763$ to $0.549$ while the unmatched rate rises from $0.076$ to $0.203$ (Table~\ref{tab:ablations}). The observer's controls (eleven buttons, i.e.\ the thirteen read by the state model except the scoreboard and use keys, plus the pitch and yaw deltas) and held weapon (a 52-way embedding) enter every DiT block through adaptive layer normalization; the other players' positions and facing directions enter only through the field.

\paragraph{Field ablations.}
Each retrained row of Table~\ref{tab:ablations} changes the field or its loss. Without camera alignment, per-player state tokens replace the field, one per latent frame and player (world position and position relative to the observer; position, bearing and log-distance in the observer's camera frame; facing direction; team; alive; presence and observer flags), appended to the cross-attention context and pooled into the latent frame's action modulation. Lacking an image position, these tokens have no visibility gate and, unlike the baseline tokens of Table~\ref{tab:main}, no visibility label; the loss keeps its foreground weight, computed outside the generator by projecting the players' positions through the observer's camera. Without visibility, every living player in the view frustum is written, occluded or not, with no visibility label. Without foreground weight, $\lambda=1$, so $\alpha_k=1$. Late injection writes the field after the 23rd DiT block; the coarse field is max-pooled by two and upsampled back. The two inference-time rows keep the \method{} weights and change one input: \emph{Field off at inference} discards the stem's entire output, biases included; \emph{Visibility off at inference} marks every player visible, so every living player other than the observer in the view frustum is written at full confidence, as by the model trained without visibility.

Table~\ref{tab:ablations-ci} reports each row of Table~\ref{tab:ablations} minus \method{}, paired on the same views, with $95\%$ bootstrap intervals over the ten-second windows. Every retrained row has a lower render rate, by $0.03$ to $0.55$ in the median per-view difference, with all intervals below zero. Visibility off at inference raises the unmatched rate by $0.14$ ($[0.11, 0.17]$), with a render-rate difference in $[-0.017, 0]$. Training without camera alignment and both inference-time changes lower PSNR and SSIM and raise LPIPS; training without visibility, without foreground weight or with a coarse field raises PSNR by $0.13$ to $0.17$\,dB; for late injection the PSNR and LPIPS intervals include zero.

\begin{table}[!htbp]
\caption{Each row of Table~\ref{tab:ablations} minus \method{}, paired over the views both rows score (PSNR, SSIM, LPIPS: mean difference; Rendered, Unmatched: median per-view difference, which need not equal the difference between the medians of Table~\ref{tab:ablations}); $^\dagger$ marks a $95\%$ interval from 10{,}000 bootstrap resamples of the windows, the two views of a window drawn together, that includes zero. Negative $\Delta$PSNR, $\Delta$SSIM and $\Delta$Rendered and positive $\Delta$LPIPS and $\Delta$Unmatched mean that the row is behind \method{}. FVD compares two distributions of clips and has no paired difference.}
\label{tab:ablations-ci}
\centering
\footnotesize
\setlength{\tabcolsep}{5pt}
\begin{tabular*}{\linewidth}{@{\extracolsep{\fill}}lccccc@{}}
\toprule
Row minus \method{} & $\Delta$PSNR & $\Delta$SSIM & $\Delta$LPIPS & $\Delta$Rendered & $\Delta$Unmatched \\
\midrule
Trained without camera alignment & $-0.37$ & $-0.005$ & $+0.024$ & $-0.545$ & $+0.560$ \\
Trained without visibility & $+0.13$ & $+0.003$ & $-0.006$ & $-0.047$ & $0.000^\dagger$ \\
Trained without foreground weight & $+0.17$ & $+0.006$ & $-0.009$ & $-0.043$ & $0.000^\dagger$ \\
Trained with late injection & $-0.02^\dagger$ & $+0.004$ & $+0.001^\dagger$ & $-0.148$ & $+0.051$ \\
Trained with a coarse field & $+0.15$ & $+0.003$ & $-0.010$ & $-0.034$ & $0.000^\dagger$ \\
Field off at inference & $-0.20$ & $-0.003$ & $+0.012$ & $-0.690$ & $+0.192$ \\
Visibility off at inference & $-0.38$ & $-0.009$ & $+0.020$ & $0.000^\dagger$ & $+0.143$ \\
\bottomrule
\end{tabular*}
\end{table}

\FloatBarrier
\section{Limitations and scope}
\label{app:lim}
This appendix states the measurement conditions of \sect{sec:experiments} and what they leave untested.

\paragraph{Player states.} The whole-round results of Tables~\ref{tab:scene}, \ref{tab:multiview}, \ref{tab:crowding}, \ref{tab:scene-whole} and~\ref{tab:scene-intervals} and of the upper part of Table~\ref{tab:scene-agree} give every client the GT states of all players, to measure the generator and the scene state separately from the predicted states (Appendix~\ref{app:longhorizon}). Table~\ref{tab:longhorizon}a and the lower part of Table~\ref{tab:scene-agree} use closed-loop deployment, with predicted states. With predicted states (Tables~\ref{tab:predicted} and~\ref{tab:longhorizon}a), the entity criterion scores each generated frame against the positions and visibility given to the generator, that is, whether each player is drawn where that player's own client has placed it. Every rollout departs from the recording as its errors accumulate, so the criterion asks whether the clients agree on one world, kept shared by the predicted states and, in Table~\ref{tab:longhorizon}a, the scene state, rather than whether that world follows the recording. On video generated with GT states, the state model's median position error against the GT trajectory grows to $2.64$ body widths at $50$--$60$\,s, and $39\%$ of clients are within two body widths at $60$\,s (Table~\ref{tab:state}).

\paragraph{The entity criterion.} Rendered counts a player as drawn when the fine-tuned detector finds a person within one body width of its position in the given player state; Unmatched counts detections with no such player nearby and no person detected nearby in the recorded video. Neither checks the team, identity, facing direction, weapon or controls that the field also carries; both inherit the detector's errors, and only players that the detector finds in the recorded video are scored for Rendered, so those it misses there are not evaluated (\sect{sec:criterion}).

\paragraph{Scene state.} Relative to the model trained without scene state, the gain in PSNR, SSIM and LPIPS holds after 30\,s but not in the first 30\,s (Appendix~\ref{app:scene}). After 30\,s, the depths predicted for two clients' generated frames at the same instant, reprojected into one camera, agree within the larger of $24$\,u and $5\%$ on $0.206$ of pixels for \method{}, against $0.365$ when the same measure is applied to the recorded video (Table~\ref{tab:scene-agree}).

\paragraph{Number of players.} The generator's computation is independent of $P$, but the scene state holds up to $B$ memory entries per client, so the memory entries projected at retrieval grow with $P$ (Appendix~\ref{app:scene}); the frame rates of Table~\ref{tab:distributed} include this cost. Where players overlap in the image, each token of the field describes at most two players, and the headcount channel saturates at four (Appendix~\ref{app:field}); since no frame of the 32 rounds has more than four other players in view with a box of at least $24^2$ pixels (Table~\ref{tab:crowding}), denser crowds are not evaluated.

\paragraph{Clients and the network.} In our evaluations, clients advance in lockstep, one block at a time (\sect{sec:framework}); the evaluations and timing measurements (Appendix~\ref{app:deploy}) vary neither network delay, jitter nor loss, and do not study clients that fall behind. Within a block, a client extrapolates the other players' latest published positions from their controls (Appendix~\ref{app:state}); the effect of staleness on rendering is not measured. Player states and memory entries from other clients are used as received, without cross-checking, so a faulty client alters what the others render.

\paragraph{What comes from the recording.} Every client starts from recorded first frames and its GT position (Appendices~\ref{app:state} and~\ref{app:longhorizon}). In the whole rounds, three players of one team run clients, and all five in Table~\ref{tab:longhorizon}a, while the other players follow their GT states; in Table~\ref{tab:predicted}, players without a recorded first-person view of the window keep their recorded states; each one-hour rollout follows one player's recorded controls and restarts its position estimate at that player's recorded spawn position in every round (Appendix~\ref{app:longhorizon}). The state model estimates only position and depth; the other attributes of a player state come from the player's own inputs (\sect{sec:field}). No component of \method{} determines hits, damage or deaths.

\paragraph{Maps and domain.} The place head classifies positions into cells of the training maps and reads a map embedding (Appendix~\ref{app:state}), so a new map requires recorded positions for training; the depth head is supervised with depth rendered from the public 3D map assets. All results are on one game and its training maps.

\paragraph{Baselines and runs.} Solaris and frame concatenation in Table~\ref{tab:main} are those designs' attention patterns in our stage-2 model, retrained with our data and budget; they keep consistency implicit in cross-view attention and receive the player states as tokens so that they know where the other players are, as the field does (Appendix~\ref{app:scaling}). The released Solaris and Gamma-World are timed with random weights (Appendix~\ref{app:deploy}) but not scored. Each model of Table~\ref{tab:training} and each retrained baseline is trained once, and no table reports variation across training runs; the bootstrap intervals resample windows or rounds.

\FloatBarrier
\section{Additional qualitative comparisons}
\label{app:qualitative}
\makeatletter
\setlength{\@fptop}{0pt}
\setlength{\@fpsep}{12pt}
\setlength{\floatsep}{6pt plus 2pt minus 2pt}
\setlength{\textfloatsep}{6pt plus 2pt minus 3pt}
\setlength{\intextsep}{5pt plus 2pt minus 2pt}
\makeatother

\begin{figure}[htbp]
\centering
\includegraphics[trim=17.75bp 23bp 15.25bp 0.5bp,clip,width=\linewidth]{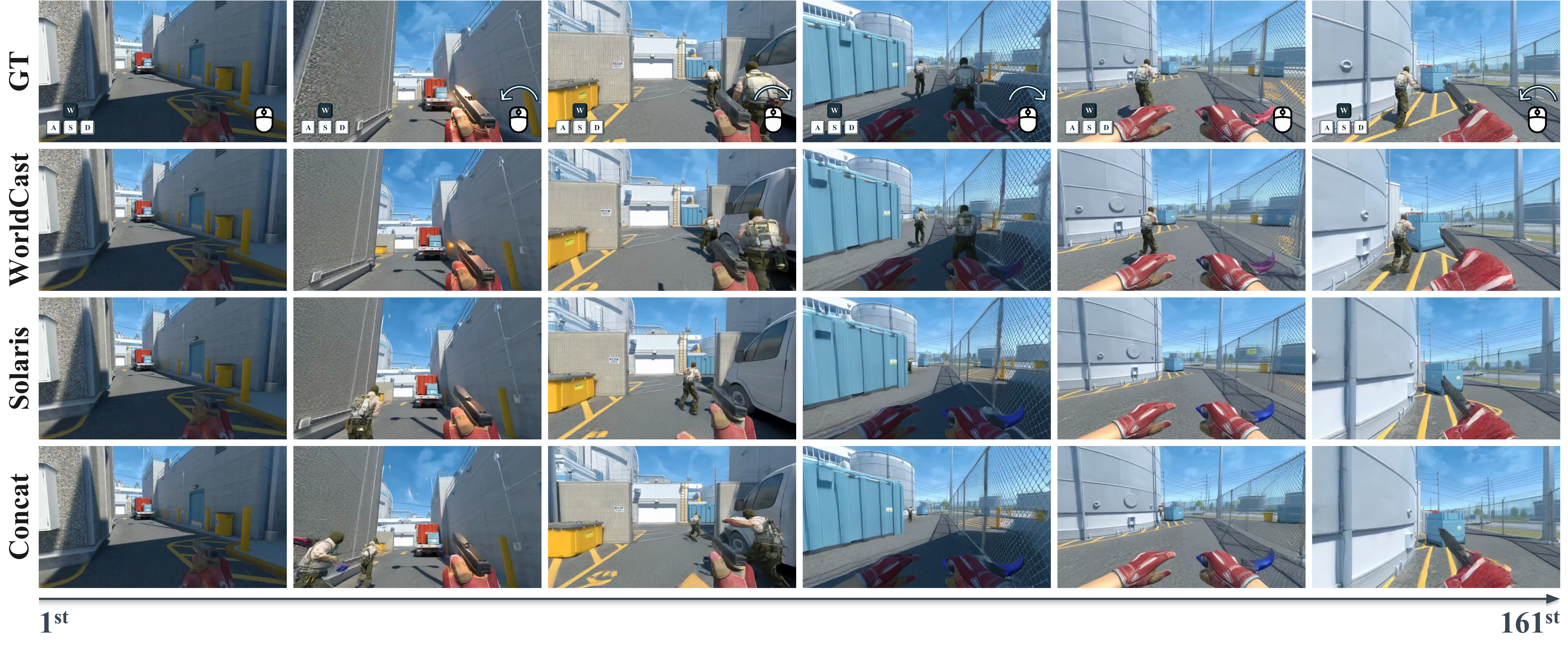}
\caption{Nuke round 13, player 2; rows and overlays as in Fig.~\ref{fig:qualitative}.}
\label{fig:qual-app1}
\end{figure}

\begin{figure}[htbp]
\centering
\includegraphics[trim=17.75bp 23bp 15.25bp 0.5bp,clip,width=\linewidth]{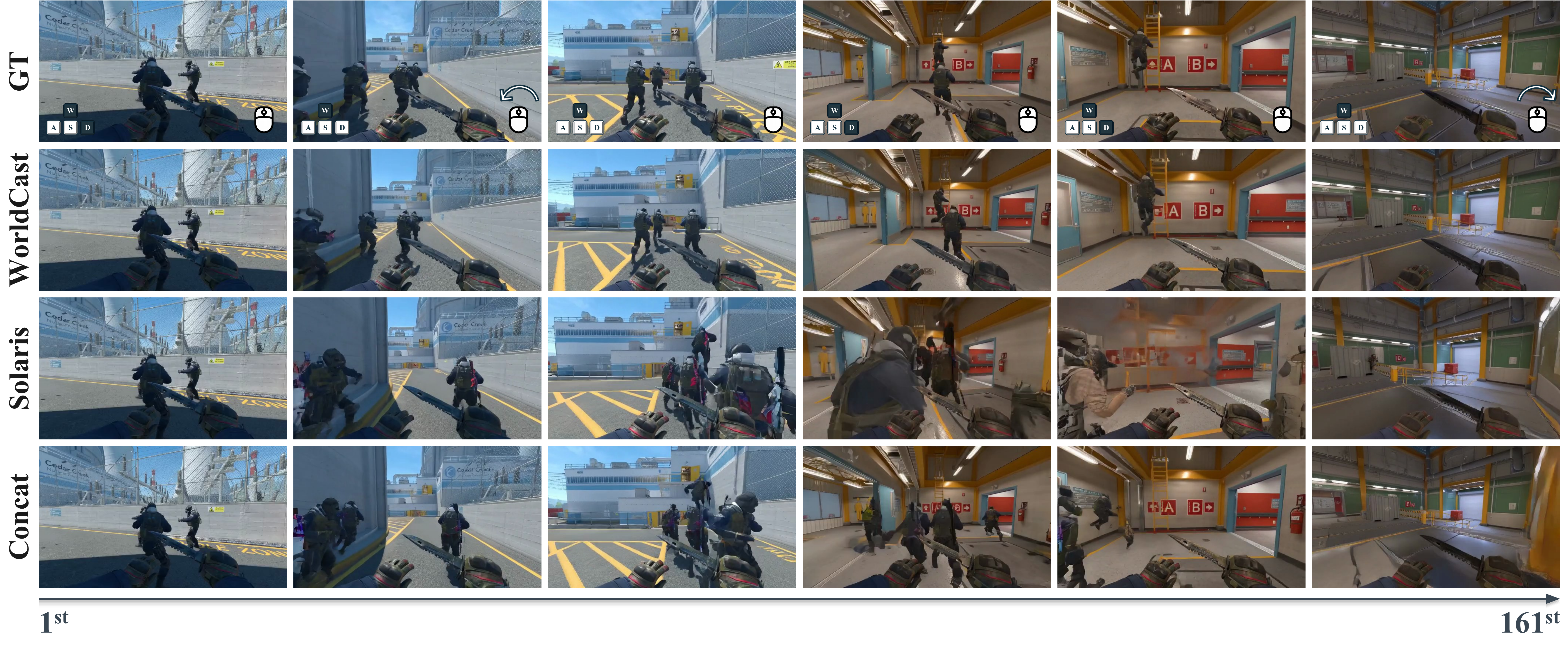}
\caption{Nuke round 13 of another match, player 1; rows and overlays as in Fig.~\ref{fig:qualitative}.}
\label{fig:qual-app2}
\end{figure}

\begin{figure}[htbp]
\centering
\includegraphics[trim=17.75bp 23bp 15.25bp 0.5bp,clip,width=\linewidth]{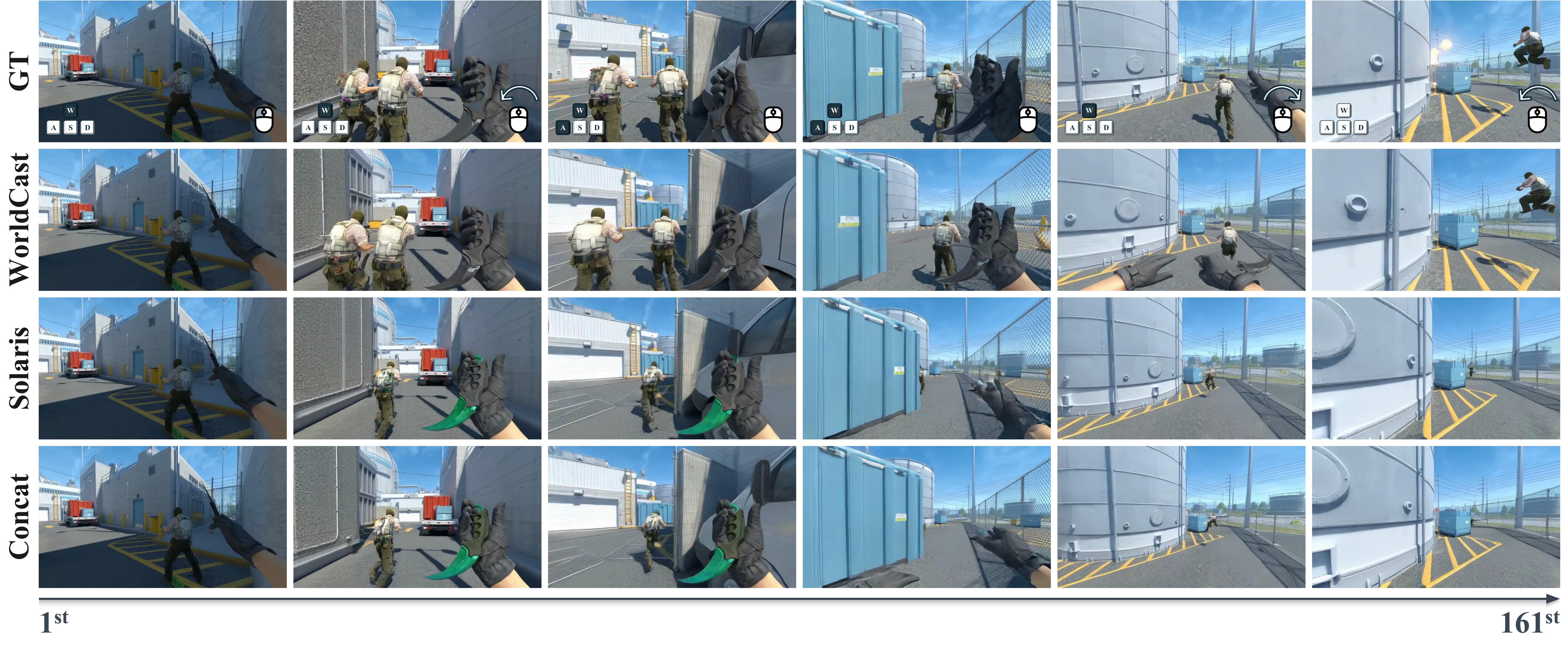}
\caption{Nuke round 2, player 7; rows and overlays as in Fig.~\ref{fig:qualitative}.}
\label{fig:qual-app3}
\end{figure}

\begin{figure}[htbp]
\centering
\includegraphics[trim=17.75bp 23bp 15.25bp 0.5bp,clip,width=\linewidth]{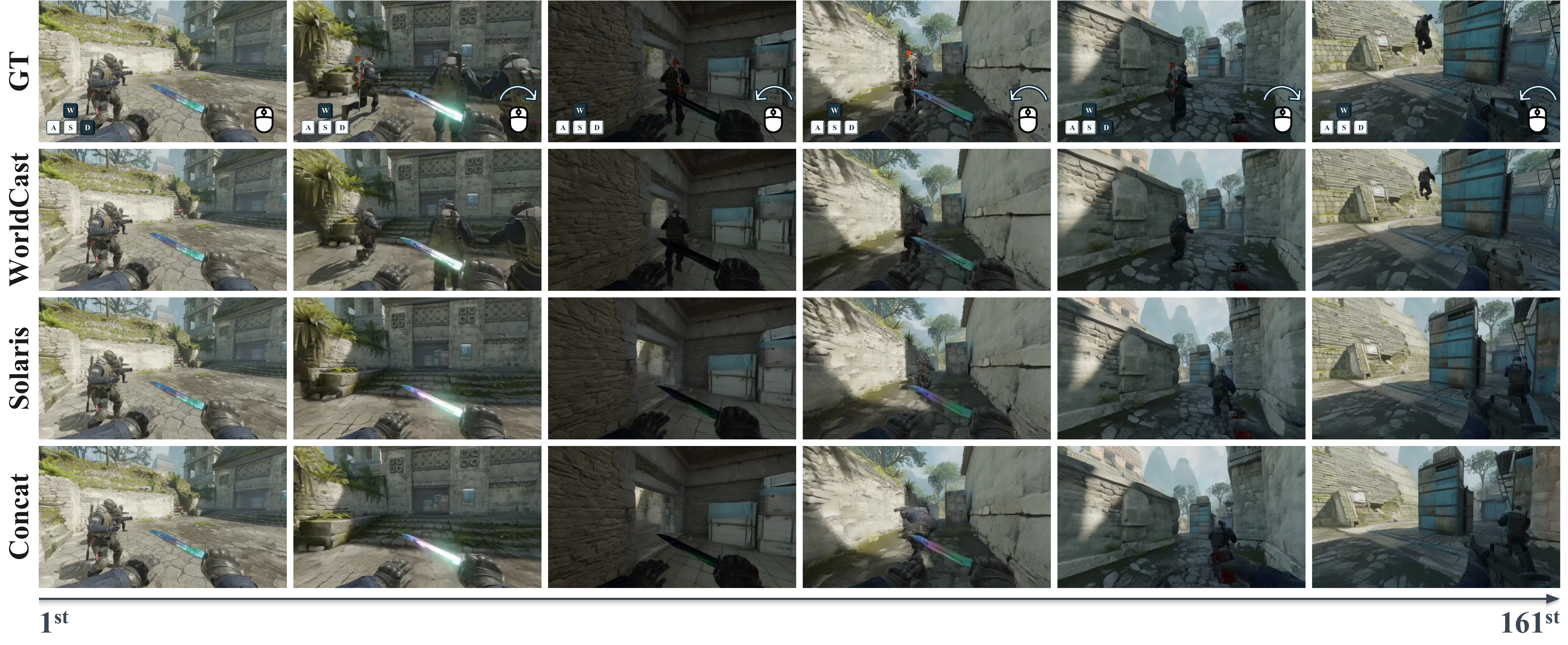}
\caption{Ancient round 22, player 5; rows and overlays as in Fig.~\ref{fig:qualitative}.}
\label{fig:qual-app4}
\end{figure}

\begin{figure}[htbp]
\centering
\includegraphics[trim=17.75bp 23bp 15.25bp 0.5bp,clip,width=\linewidth]{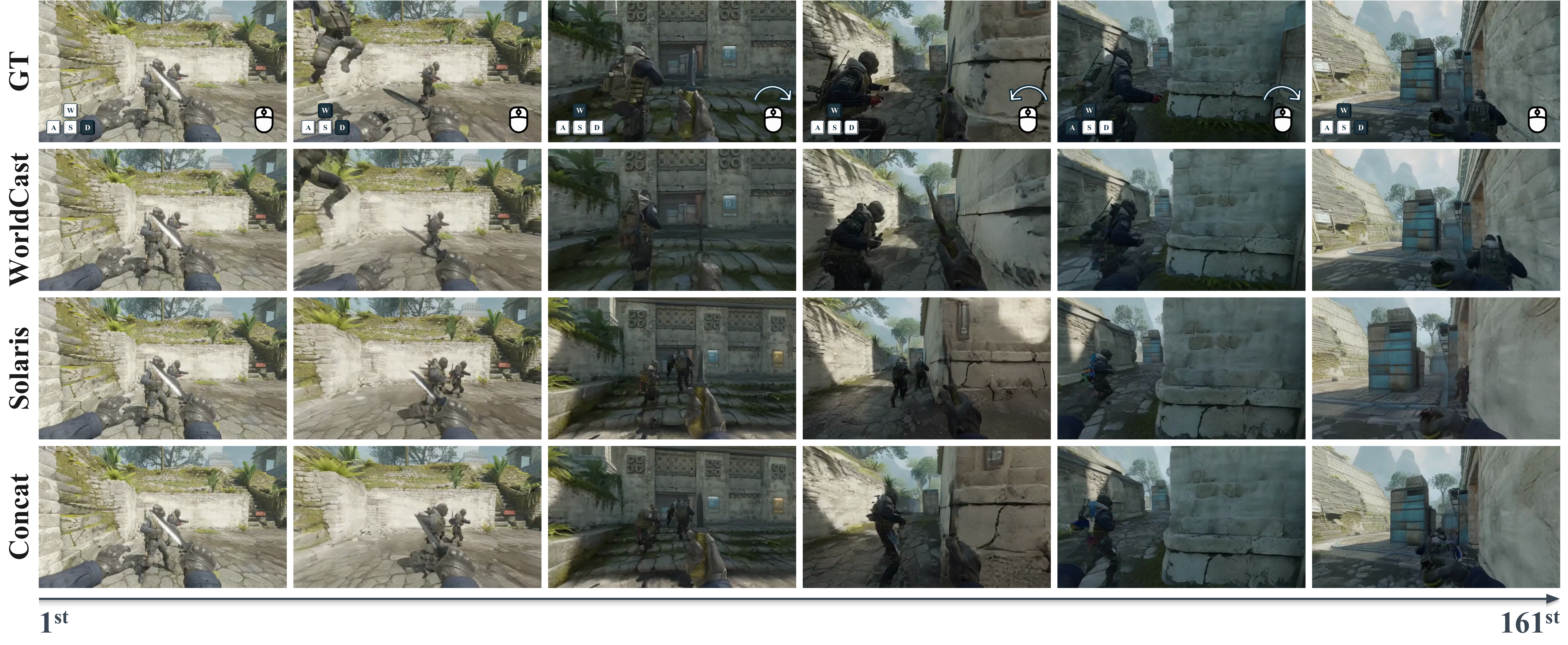}
\caption{Ancient round 18, player 1; rows and overlays as in Fig.~\ref{fig:qualitative}.}
\label{fig:qual-app5}
\end{figure}

\begin{figure}[p]
\centering
\includegraphics[trim=17.75bp 23bp 15.25bp 0.5bp,clip,width=\linewidth]{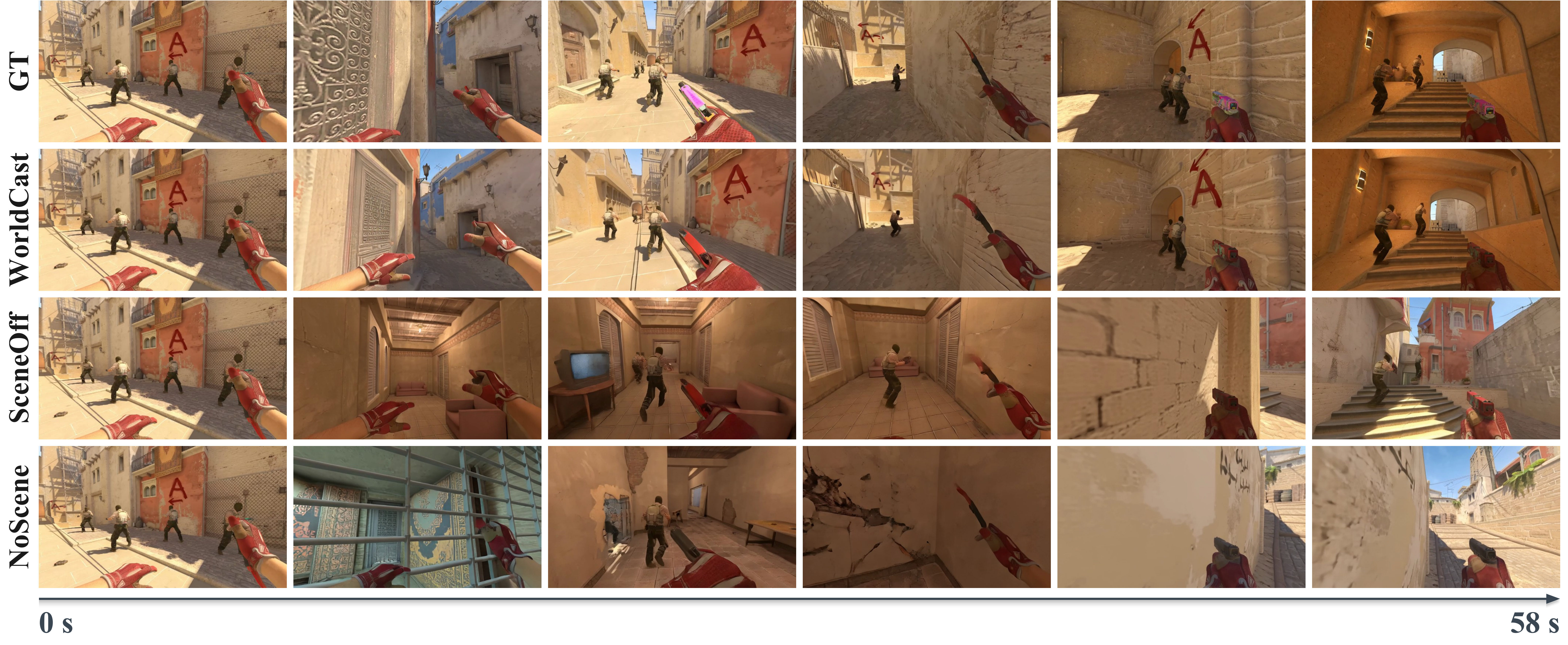}
\caption{Mirage round 16, one client, whole round with GT states. GT: recording; SceneOff: scene state off at inference; NoScene: trained without scene state. After 30\,s, \method{} keeps the scene of the recording and in the third column returns to the wall of its first frame; SceneOff and NoScene generate other parts of the map.}
\label{fig:scene-mirage}
\par\vspace{12pt}
\includegraphics[trim=17.75bp 23bp 15.25bp 0.5bp,clip,width=\linewidth]{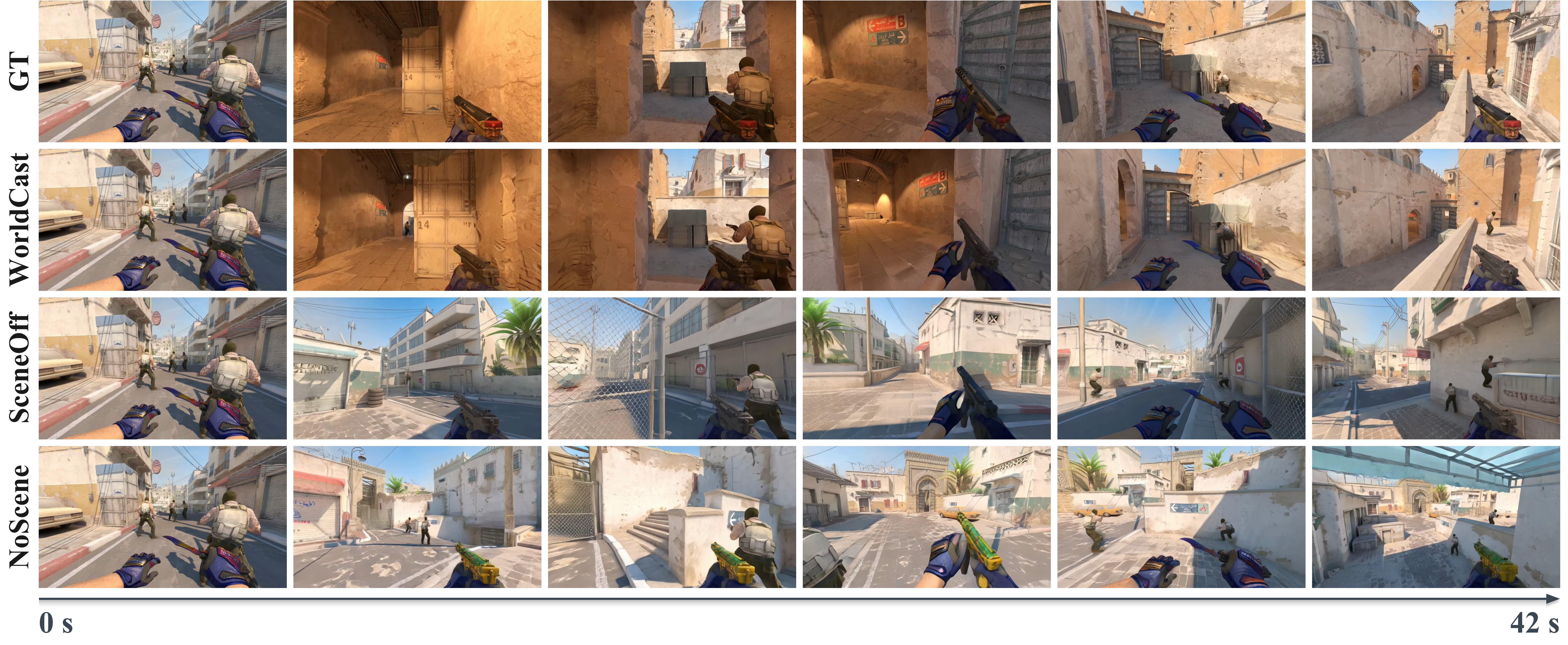}
\caption{Dust2 round 9, one client; rows as in Fig.~\ref{fig:scene-mirage}.}
\label{fig:scene-dust2}
\end{figure}

\begin{figure}[p]
\centering
\includegraphics[trim=17.75bp 23bp 15.25bp 0.5bp,clip,width=\linewidth]{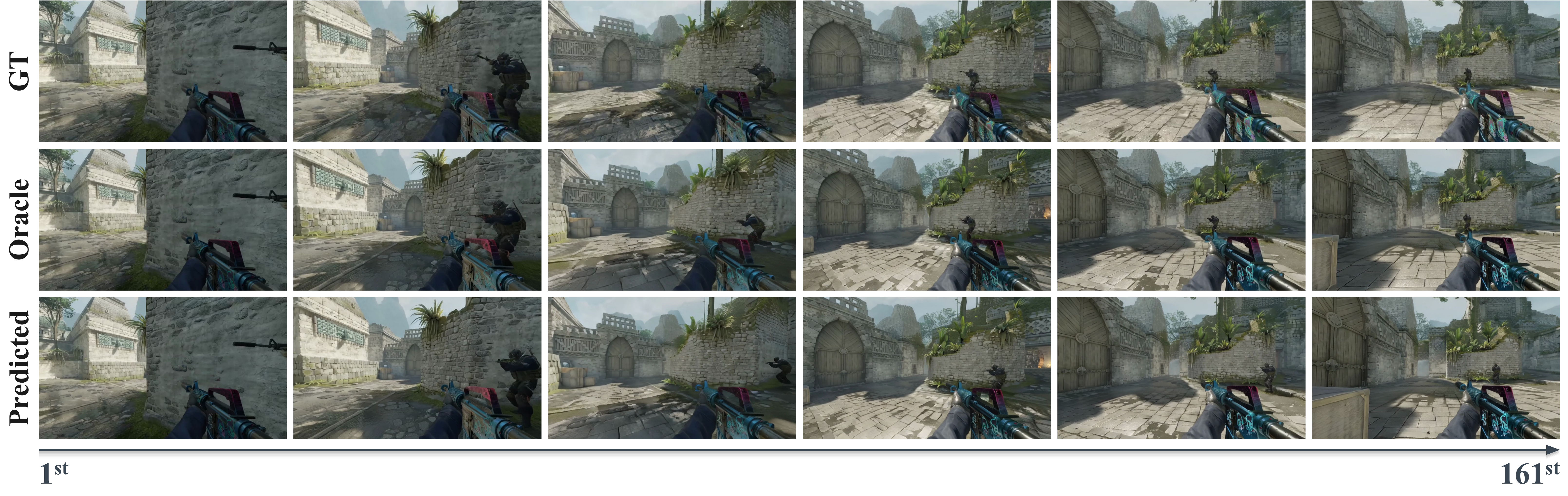}
\caption{Ancient round 11, one player, ten-second window of Table~\ref{tab:predicted}. GT: recording; Oracle: GT positions; Predicted: positions predicted in closed loop, every client estimating its own position from its generated video, with visibility from the depth head.}
\label{fig:window-pred}
\par\vspace{12pt}
\includegraphics[trim=17.75bp 23bp 15.25bp 0.5bp,clip,width=\linewidth]{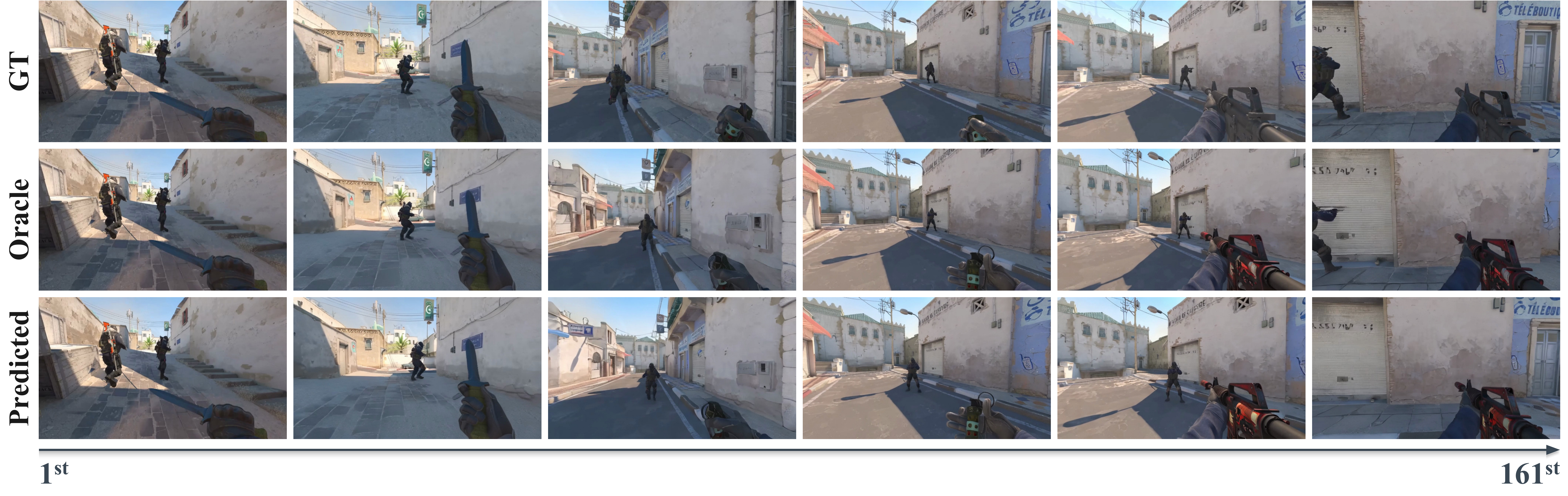}
\caption{Dust2 round 18, one player; rows as in Fig.~\ref{fig:window-pred}.}
\label{fig:window-pred-b}
\end{figure}

\begin{figure}[p]
\centering
\includegraphics[trim=17.75bp 23bp 15.25bp 0.5bp,clip,width=\linewidth]{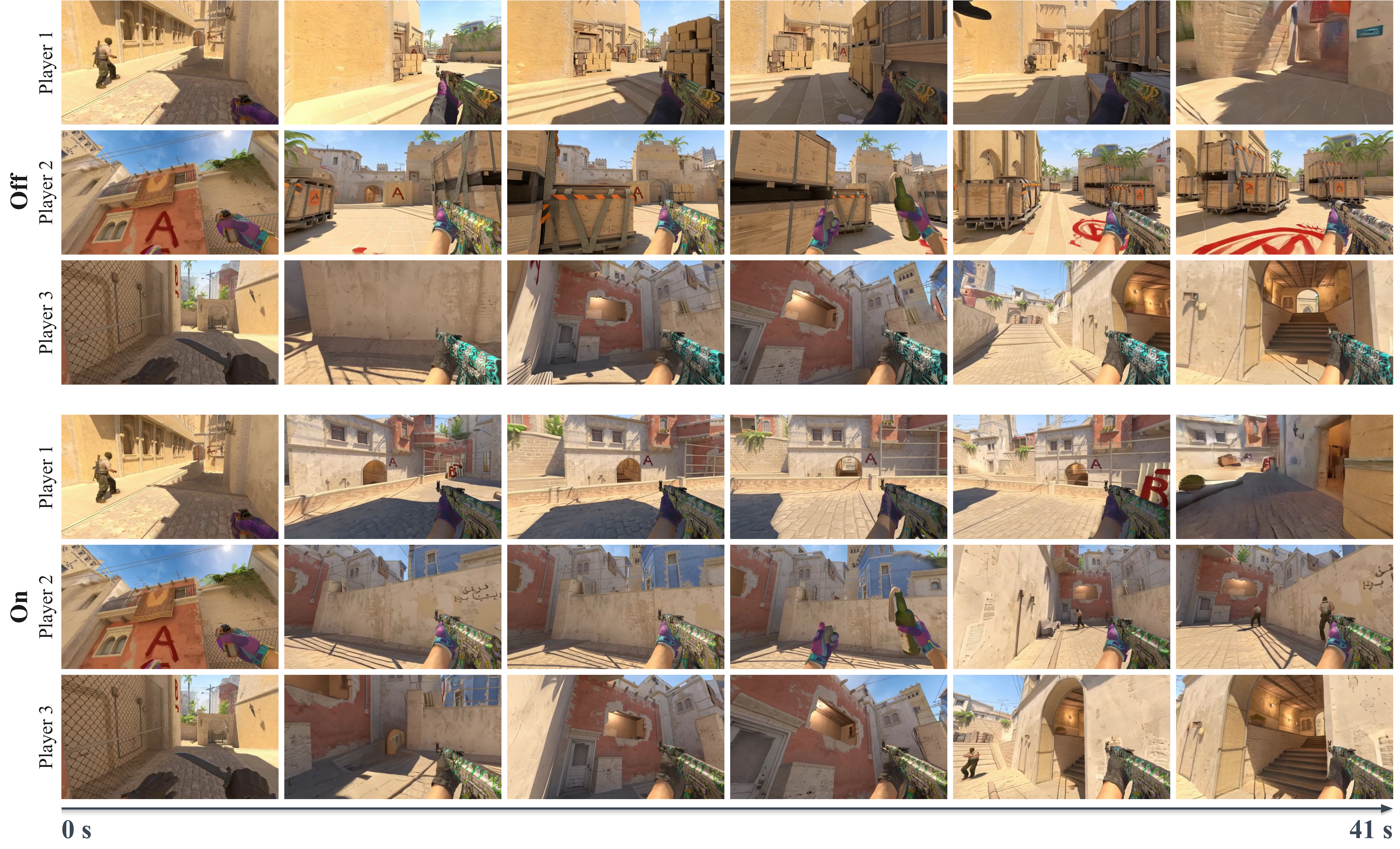}
\caption{Mirage round 11, closed-loop deployment of Table~\ref{tab:longhorizon}a: all five players of one team run clients with predicted states; three are shown. Off/On: scene state off/on; a player's Off and On rows differ only in scene state. With scene state, the three views stay around the A-site ramp and agree on its landmarks, and Players 2 and 3 draw the teammates ahead of them (supplementary videos).}
\label{fig:cl-mirage}
\par\vspace{12pt}
\includegraphics[trim=17.75bp 23bp 15.25bp 0.5bp,clip,width=\linewidth]{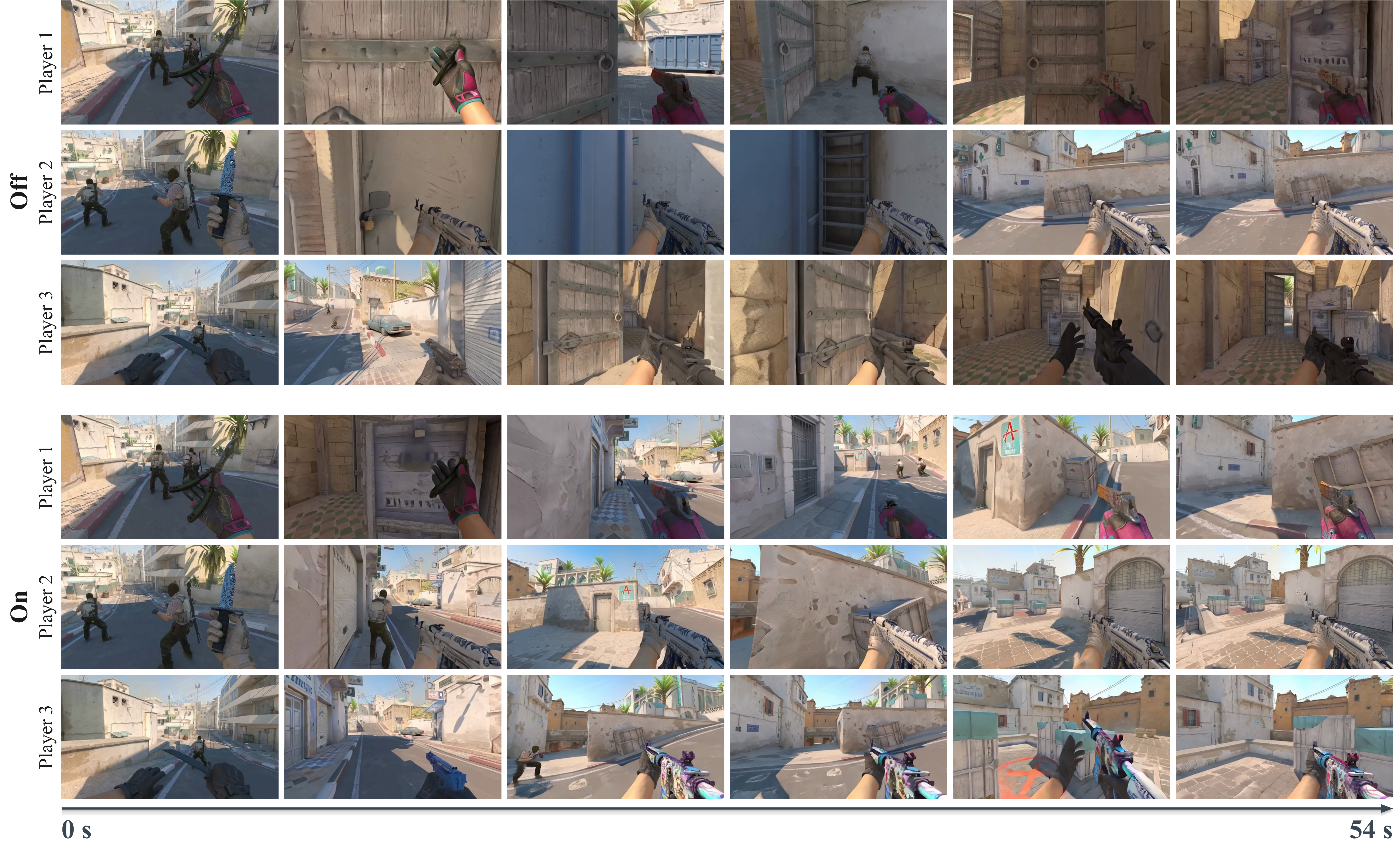}
\caption{Dust2 round 22; rows as in Fig.~\ref{fig:cl-mirage}. With scene state, the three players push from mid to A site and see their teammates ahead; without it, Players 1 and 3 remain in a corner by a wooden door and Player 2 faces a wall.}
\label{fig:cl-dust2}
\end{figure}

Figs.~\ref{fig:qual-app1}--\ref{fig:qual-app5} show five further held-out rounds, each from a different match and from the start of the round, in the layout of Fig.~\ref{fig:qualitative}: the recording with its controls overlaid, \method{}, Solaris and frame concatenation at the same six frames. In the same six-column layout, Figs.~\ref{fig:scene-mirage} and~\ref{fig:scene-dust2} show whole rounds with and without scene state (Table~\ref{tab:scene}), and Figs.~\ref{fig:window-pred} and~\ref{fig:window-pred-b} ten-second windows with GT and predicted states (Table~\ref{tab:predicted}), and Figs.~\ref{fig:cl-mirage} and~\ref{fig:cl-dust2} closed-loop deployment with scene state on and off (Table~\ref{tab:longhorizon}a); all rows start from the same recorded first frame.

\end{document}